\documentclass{article} % For LaTeX2e

\usepackage{arxiv}

\usepackage{amsmath,amsfonts,bm}

\def\eqref#1{equation~\ref{#1}}
\def\1{\bm{1}}

\def\vh{{\bm{h}}}

\def\vu{{\bm{u}}}

\def\vw{{\bm{w}}}
\def\vx{{\bm{x}}}

\def\mA{{\bm{A}}}

\def\mC{{\bm{C}}}
\def\mD{{\bm{D}}}

\def\mH{{\bm{H}}}
\def\mI{{\bm{I}}}

\def\mP{{\bm{P}}}
\def\mQ{{\bm{Q}}}

\def\mS{{\bm{S}}}

\def\mU{{\bm{U}}}

\def\mX{{\bm{X}}}
\def\mY{{\bm{Y}}}
\def\mZ{{\bm{Z}}}

\DeclareMathAlphabet{\mathsfit}{\encodingdefault}{\sfdefault}{m}{sl}
\SetMathAlphabet{\mathsfit}{bold}{\encodingdefault}{\sfdefault}{bx}{n}

\usepackage{hyperref}
\usepackage{url}

\usepackage[acronym, nohypertypes={acronym}]{glossaries}
\glsdisablehyper

\newacronym{higp}{HiGP}{Hierarchical Graph Predictor}
\newacronym{shr}{SHR}{\textit{Stochastic Hierarchical Regularization}}
\newacronym{fr}{FR}{\textit{forecast reconciliation}}

\newacronym{stgnn}{STGNN}{spatiotemporal graph neural network}
\newacronym{gnn}{GNN}{graph neural network}
\newacronym{mtl}{MTL}{multi task learning}
\newacronym{mp}{MP}{message-passing}
\newacronym{tts}{TTS}{time-then-space}
\newacronym{mlp}{MLP}{multi-layer perceptron}
\newacronym{rnn}{RNN}{recurrent neural network}
\newacronym{sota}{SOTA}{state of the art}
\newacronym{dcrnn}{DCRNN}{diffusion convolution recurrent neural network}
\newacronym{src}{SRC}{\textit{select, reduce, connect}}
\newacronym{bu}{BU}{\textit{bottom-up}}
\newacronym{td}{TD}{\textit{top-down}}
\newacronym{fm}{FM}{foundation model}
\newacronym{dl}{DL}{deep learning}
\newacronym{mae}{MAE}{Mean Absolute Error}
\newacronym{kmis}{k-MIS}{$k$-maximal independent set}

\newglossaryentry{la}{name=Metr-LA,description=}
\newglossaryentry{bay}{name=PeMS-Bay,description=}
\newglossaryentry{cer}{name=CER-E,description=}
\newglossaryentry{air}{name=AQI,description=}
\newglossaryentry{largest}{name=LargeST,description=}
\newglossaryentry{era5}{name=ERA5,description=}
\newglossaryentry{gla}{name=GLA,description=}
\newglossaryentry{pems3}{name=PEMS03,description=}
\newglossaryentry{pems4}{name=PEMS04,description=}
\newglossaryentry{pems7}{name=PEMS07,description=}
\newglossaryentry{pems8}{name=PEMS08,description=}
\newglossaryentry{engrad}{name=EngRAD,description=}

\newglossaryentry{ttsimp}{name=TTS-IMP,description=}
\newglossaryentry{ttsamp}{name=TTS-AMP,description=}
\newglossaryentry{tasimp}{name=T\&S-IMP,description=}
\newglossaryentry{tasamp}{name=T\&S-AMP,description=}

\usepackage{enumitem}
\usepackage{natbib}
\usepackage{graphicx}
\usepackage{booktabs}
\usepackage{subcaption}
\usepackage{wrapfig}
\usepackage[normalem]{ulem}

\title{Learning to Forecast on Multiple Spatiotemporal Hierarchies}

\author{
Valentina Moretti$^{1}$\thanks{Correspondence to: \texttt{valentina.moretti@usi.ch}}
\quad
Cesare Alippi$^{1,2}$ \quad
Andrea Cini$^{3}$\\[0.5em]
$^{1}$IDSIA USI-SUPSI, Universit\`a della Svizzera italiana\\
$^{2}$Politecnico di Milano\\
$^{3}$EPFL
}

\begin{document}

\maketitle

\begin{abstract}

Hierarchical forecasting exploits aggregation structures that expose multiple views of the same underlying spatiotemporal forecasting problem, while ensuring coherence among forecasts. 
Individual series provide fine-grained information, while progressively coarser aggregates reveal temporal and spatial patterns at different scales. 
In the associated deep learning architectures, the hierarchy determines which aggregate series are available and how they are processed together with individual series~(e.g., through reconciliation or message-passing operators). 
Different hierarchies thus provide distinct learning signals and inductive biases for forecasting the same collection of time series. 
Nonetheless, existing architectures are typically trained on a single, often fixed, structure.
In this work, we propose a graph deep learning framework that leverages variations across hierarchies as multi-task regularization.
We show that jointly learning across these structures can reduce over-specialization to any single hierarchy and improve forecasting performance. 
Indeed, training across hierarchies provides an inductive bias toward representations that are useful across different structures, and as such, generalize better.
Making forecasts coherent with the aggregation constraints for each sampled hierarchy is computationally expensive: to complement the framework, we propose a novel operator---based on a top-down approach---that produces coherent forecasts with linear time and space complexity w.r.t.\ the hierarchy size. Experiments show that our approach improves forecasting accuracy over training on a single structure and is competitive with current state-of-the-art methods.

\end{abstract}

\section{Introduction}
\label{s:intro}

Many forecasting problems involve collections of time series recorded by sensors at different locations. Spatiotemporal forecasting addresses these settings by jointly modeling the temporal dynamics of individual series and spatial dependencies among them. Recent deep learning architectures represent the irregular structure of the problem by using graphs, where sensors are nodes and edges encode pairwise relationships among them~(e.g., spatial proximity)~\citep{jin_2024_survey_gnn,cini_2025_graph_deep}. 
Recently, hierarchical representations have been increasingly used in this context~\citep{yu_2021_stunet,hermes_2022_unet_traffic,chen_2023_gnn_air_quality}. 
A hierarchy specifies how input series are progressively grouped into hierarchical clusters, producing aggregates at different resolutions that can highlight relevant dynamics in the underlying process. Hierarchical representations also model relationships beyond pairwise interactions~(graph edges), allowing information to propagate across levels---e.g., with appropriate \gls{mp} operators~\citep{cini_2024_higp}.
The hierarchical structure additionally creates shorter paths between distant nodes, allowing information to travel across larger portions of the graph within a limited number of \gls{mp} layers~\citep{cini_2024_higp,carotti_2026_graph_hierarchical_recurrence}.
As such, hierarchical models have found several practical applications in spatiotemporal forecasting, including pandemic~\citep{ma_2022_hierarchical_pandemic}, air-quality~\citep{chen_2023_gnn_air_quality}, traffic~\citep{guo_2021_hierarchical,hermes_2022_unet_traffic}, and financial forecasting~\citep{arya_2023_hierarchical_stock}.
Forecasting aggregate series offers two further advantages. First, aggregation structures encode coherence constraints that can be used to regularize predictions through reconciliation
~(forecasts at different levels should \textit{add up})
~\citep{wickramasuriya_2019_trace_minimization,panagiotelis_2021_reconciliation_geometric,panagiotelis_2023_probabilistic_reconciliation}. 
By leveraging predictions from the entire hierarchy to enforce these constraints, reconciliation acts as a forecast combination that can reduce variance and improve accuracy~\citep{bates_1969_forecast_combination,hollyman_2021_understanding_reconciliation,difonzo_2024_combination_reconciliation}.
Second, aggregate series provide additional training targets, allowing the model to learn from errors at both individual and aggregate levels~\citep{rangapuram_2021_endtoend,cini_2024_higp}. 

\begin{figure}[t]
  \centering
    \includegraphics[width=1.0\linewidth]{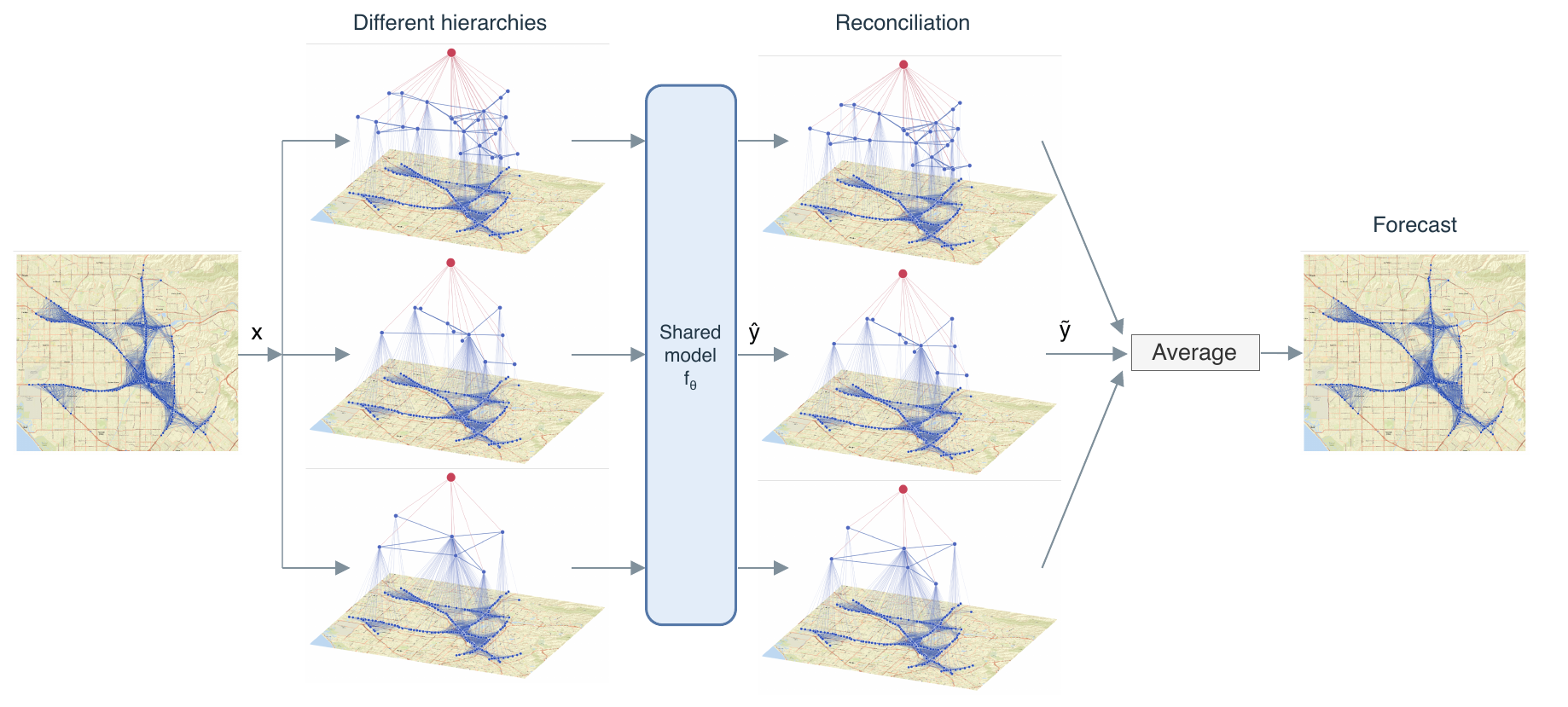}
  \caption{Inference step for a model trained with our framework on \gls{largest}. Each input is processed and reconciled under multiple hierarchies, and the resulting bottom-level forecasts are averaged.}
  \label{fig:multi_hierarchy_pipeline_largest}
\end{figure}
Taken together, these aspects make the hierarchy an integral part of the forecasting model and a source of inductive bias: it determines the aggregate series available to the model, the paths through which information propagates across scales, the higher-level targets used for supervision, and the constraints imposed on the forecasts.  
Consequently, different hierarchies group the series into different aggregates and connect them through different paths, thereby inducing distinct but related learning tasks that share the same bottom-level forecasting objective. 
Related tasks can encourage the model to learn representations that are useful across different structures and generalize better to the common forecasting task, as established in the multi-task learning literature~\citep{caruana_1993_multitask}. This effect arises when parameters are shared across tasks, favoring solutions useful across them over those specialized to a single one.
However, existing deep hierarchical forecasting methods generally train on a single, often fixed, hierarchy~\citep{rangapuram_2021_endtoend,yang_2017_reconciling,cini_2024_higp}.

In this paper, we introduce a framework that leverages multiple, structurally diverse hierarchies as a source of regularization for spatiotemporal forecasting. In particular, we train a single hierarchical forecasting model over a distribution of hierarchies, sampling a different structure at each training step. Sharing parameters across different hierarchies encourages the same model's forecasting and hierarchical \gls{mp} functions to perform well across different groupings, thereby reducing overfitting to any single hierarchy. 
This may be particularly useful in inductive settings, where a trained model is applied to previously unseen time series. 
Training over different hierarchies, however, introduces a computational challenge, as widely used reconciliation methods require dense matrix inversions for each sampled structure and scale poorly with hierarchy size~\citep{hyndman_2016_fast_reconciled,sprangers_2024_forecasting_at_scale,petropoulos_2025_scalable_sub_hierarchies,sharma_2025_hierarchical_network_flow}. To tackle the problem, we introduce a recursive operator that combines forecasts across adjacent hierarchy levels, ensuring coherence by construction with linear time and space complexity w.r.t.\ the hierarchy size.
The main contributions of this paper are:
\begin{itemize}[topsep=.1em, leftmargin=1em, itemindent=1em]
    \item We introduce a training framework that optimizes a shared spatiotemporal forecasting model across multiple structurally diverse hierarchies, using their variation as a source of multi-task regularization. Learning across tasks leads to more effective and general representations.
    \item We develop an efficient recursive procedure that incorporates forecasts from every aggregation level while enforcing the corresponding coherence constraints, making training across a distribution of hierarchies computationally practical.
    \item We instantiate the proposed framework in a spatiotemporal graph architecture that combines \glspl{gnn}, graph pooling, and hierarchical information propagation. This implementation allows the same model parameters to be trained across different hierarchies while preserving coherence w.r.t.\ each sampled structure.
\end{itemize}
We evaluate the framework across spatiotemporal forecasting benchmarks, aggregation structures, and inductive settings. 
Experiments show that jointly training across hierarchies improves accuracy compared with training on individual hierarchies, supporting the hypothesis that variation in the aggregation structures can act as a source of regularization for the forecasting task.

\section{Preliminaries}
\label{prel:preliminaries}
\paragraph{Spatiotemporal forecasting.}
Let $\mX_t \in \mathbb{R}^{N \times d}$ denote the values of $N$ multivariate time series at time $t$; $\vx_t^i \in \mathbb{R}^{d}$ is the $d$-dimensional observation associated with node $i$. The spatial relations among time series are represented by a weighted graph with adjacency
matrix $\mA \in \mathbb{R}^{N \times N}$, where $a_{ij}$ is the weight of the edge from node $i$ to node $j$. We denote by $\mU_t \in \mathbb{R}^{N \times d_u}$ the exogenous covariates.
Given a history $\mX_{t-W+1:t}$ of $W$ steps in the range $[t-W+1,t]$, the model $g_\theta$ produces point forecasts for the next $H$ observations as $\widehat{\mX}_{t+1:t+H} = g_\theta\!\left(\mX_{t-W+1:t}, \mU_{t-W+1:t}, \mA\right)$. 
We use \gls{tts} networks that first encode each series' recent history and then exchange information among connected nodes via \gls{mp} \citep{gao_2021_equivalence,cini_2025_graph_deep}.
For node $i$, a generic layer $k$ of such \gls{stgnn} can be written as
\begin{align}
    \vh_t^{i,0}
    &= \mathrm{SeqEnc}\!\left(
        \vx_{t-W+1:t}^i,\,
        \vu_{t-W+1:t}^i
    \right), 
    \label{eq:seq-enc}
    \\
    \vh_t^{i,k+1}
    &= \mathrm{Update}^{k}\!\left(
        \vh_t^{i,k},
        \mathrm{Aggr}_{j\in\mathcal{N}(i)}
        \left\{
            \mathrm{Msg}^{k}
            \left(\vh_t^{i,k},\vh_t^{j,k},a_{ij}\right)
        \right\}
    \right),
\end{align}
where $\mathrm{Update}^k({}\cdot{})$ and $\mathrm{Msg}^k({}\cdot{})$ denote update and message functions, $\mathrm{SeqEnc}({}\cdot{})$ the network performing temporal encoding for each input series, and $\mathrm{Aggr}\{{}\cdot{}\}$ the permutation-invariant aggregation function over the set of neighbors $\mathcal{N}(i)$ of the $i$-th node.

\paragraph{Hierarchical forecasting.}

In hierarchical forecasting, a hierarchy organizes a collection of $N$ time series $\mX_t\in\mathbb{R}^{N\times d}$ into progressively coarser groups. The input series constitute the bottom level, denoted by $\mY_t^{(0)}=\mX_t$.
Each higher level $l\in\{1,\ldots,L-1\}$ contains $N_l$ series  $\mY_t^{(l)}\in\mathbb{R}^{N_l\times d}$. Overall, the hierarchy contains $M=\sum_{l=0}^{L-1}N_l$ series, stacked from the highest aggregation level to the bottom in  $\mY_t\in\mathbb{R}^{M\times d}$. The matrix $\mC\in\{0,1\}^{(M-N)\times N}$ specifies which bottom-level series contribute to each aggregate. Since aggregates are represented as means, we define
$\bar{\mC}=\mD^{-1}\mC$, where $\mD=\mathrm{diag}(\mC\mathbf{1})$ stores the corresponding group sizes. The aggregation matrix is therefore obtained as $\mS=[\bar{\mC}^{\top},\mI_N]^{\top}\in\mathbb{R}^{M\times N}$, where $\mI_N\in\mathbb{R}^{N\times N}$ is the identity matrix \citep{hyndman_2016_fast_reconciled}. The hierarchy and its aggregation structure satisfy:
\begin{equation}
    \mY_t=\mS\mX_t,
    \qquad
    \mQ\mY_t=
    \begin{bmatrix}
        \mI_{M-N} \mid -\bar{\mC}
    \end{bmatrix}\mY_t
    =\mathbf{0},
    \label{eq:coherency}
\end{equation}
where $\mQ$ encodes the coherence constraints.
A forecast $\widehat{\mY}_t$ is coherent if $\mQ\widehat{\mY}_t=\mathbf{0}$, that is, if each aggregate forecast equals the mean of its corresponding bottom-level forecasts. Because forecasts produced independently at the hierarchy nodes generally do not satisfy this condition, forecast reconciliation enforces the aggregation constraints on the model outputs. 

\paragraph{Reconciliation methods and scalability}

Forecast reconciliation adjusts predictions across hierarchy levels to satisfy the aggregation constraints, transforming the unreconciled forecasts $\widehat{\mY}_t$ into coherent forecasts $\overline{\mY}_t$. This mapping is generally written as $\overline{\mY}_t=\mS\mP\widehat{\mY}_t$, where $\mP\in\mathbb{R}^{N\times M}$ combines the $M$ node-level forecasts into $N$ bottom-level forecasts and $\mS$ maps them back to all hierarchy levels~\citep{wickramasuriya_2019_trace_minimization}.
Simple methods such as \gls{bu} and \gls{td} use forecasts from a single level, respectively aggregating the bottom-level forecasts or distributing the top-level forecast throughout the hierarchy. In doing so, they discard forecasts from the remaining levels, which can provide complementary information and improve accuracy
\citep{bates_1969_forecast_combination,hollyman_2021_understanding_reconciliation,difonzo_2024_combination_reconciliation}. 
This motivates methods that combine forecasts from the entire hierarchy through weighted projections onto the coherent subspace \citep{hyndman_2011_optimal,wickramasuriya_2019_trace_minimization}.
In particular, deep hierarchical models~\citep{rangapuram_2021_endtoend,cini_2024_higp} usually obtain coherent forecasts as:
\begin{equation}
    \overline{\mY}_t =
    \arg\min_{\mZ:\,\mQ\mZ=\mathbf{0}}
    \lVert\mZ-\widehat{\mY}_t\rVert_2^2
    =
    \left[\mI_M-\mQ^\top
    \left(\mQ\mQ^\top\right)^{-1}\mQ\right]\widehat{\mY}_t.
    \label{eq:projection_reconciliation}
\end{equation}
This formulation, however, requires inverting dense matrices that can become increasingly expensive as the hierarchy grows, limiting its application to large hierarchies~\citep{hyndman_2016_fast_reconciled,sprangers_2024_forecasting_at_scale,petropoulos_2025_scalable_sub_hierarchies,sharma_2025_hierarchical_network_flow}.

\section{Stochastic Hierarchy Regularization for Spatiotemporal Forecasting}
\label{sec:method}

We use a distribution of hierarchies as a source of regularization for forecasting. 
For each input graph, we define a hierarchy distribution and sample one hierarchy at each training step. 
Each sampled hierarchy groups the same bottom-level nodes differently and connects each aggregate to the nodes it groups.
Changing the hierarchy thus changes which aggregates the model processes and how information is exchanged across levels, while the model parameters remain shared.
First, in~\autoref{method:inductive}, we formulate the optimization of shared parameters across sampled hierarchies. We then discuss how to construct the hierarchy distribution in~\autoref{method:distribution}. Next, in~\autoref{method:pyramidal}, we present the pyramidal forecaster, which processes the structure induced by the sampled hierarchy and produces predictions for all of its nodes. Finally, in~\autoref{method:reconcile}, we introduce a recursive operator that enforces coherence through local operations between adjacent levels while allowing aggregate forecasts to contribute to bottom-level predictions.

\subsection{Learning over a distribution of hierarchies}
\label{method:inductive}
Let the input at time $t$ be $\mathcal{G}_t=\langle\mX_{t-W+1:t},\mU_{t-W+1:t},\mA\rangle$. We propose a forecasting framework with two key ingredients:
\begin{enumerate}[nosep,leftmargin=*]
    \item A hierarchy distribution $\mathcal P(\cdot\mid\mA)$ over different clusterings of the input graph. Each sampled hierarchy $\mathcal H$ defines aggregate series, connections across levels, aggregate targets  $\mY^{\mathcal H}_{t+1:t+H}$, and coherence constraints as defined in \autoref{prel:preliminaries}.
    \item A predictor $f_\theta(\mathcal G_t,\mathcal H)$ that can operate on a generic input graph $\mathcal G_t$ under any hierarchy sampled from $\mathcal P(\cdot\mid\mA)$ and produce forecasts for its $N_l^{\mathcal H}$ nodes at each level $l$. Its parameters $\theta$ are shared across hierarchy samples. 
\end{enumerate}
At each training step, we sample $\mathcal H\sim\mathcal P(\cdot\mid\mA)$ and apply $f_\theta(\mathcal G_t,\mathcal H)$. We evaluate its forecasts against the original series at $l=0$ and the aggregates defined by $\mathcal H$ at $l>0$. These losses yield the following objective:
\begin{equation}
\min_{\theta}\;
\mathbb{E}_{\mathcal G_t}
\mathbb{E}_{\mathcal H\sim\mathcal P(\cdot\mid\mA)}
\left[
\sum_{l=0}^{L-1}
\mathcal L\!\left(
\left[f_\theta(\mathcal G_t,\mathcal H)\right]^{(l)},
\mY^{\mathcal H,(l)}_{t+1:t+H}
\right)
\right],
\label{eq:multi-hierarchy-objective}
\end{equation}
where
$[\cdot]^{(l)}$ selects the forecasts at level $l$, and $\mathcal L$ is the loss function. 
The bottom-level targets are shared across all sampled hierarchies, whereas the aggregate targets and computational graph depend on $\mathcal H$. 
Sharing parameters across these related objectives provides multi-task regularization that discourages overfitting to any single hierarchy and, as such, can improve generalization. Importantly, this regularization arises from training shared parameters across structurally different views of the same forecasting problem, rather than from training separate hierarchy-specific models.
At inference, the same model can produce one bottom-level forecast under each sampled hierarchy. 
In practice, we obtain forecasts for the bottom level by independently sampling $R$ hierarchies from $\mathcal P(\cdot\mid\mA)$, evaluating the model $f_\theta(\mathcal G_t,\mathcal H_r)$ under each hierarchy $H_r$, and averaging the resulting bottom-level forecasts
(see \autoref{fig:multi_hierarchy_pipeline_largest}):
\begin{equation}
\widehat{\mY}^{(0)}_{t+1:t+H}
=
\frac{1}{R}
\sum_{r=1}^{R}
\left[
f_\theta\!\left(
\mathcal{G}_t,
\mathcal{H}_r
\right)
\right]^{(0)}.
\label{eq:hierarchy_averaging}
\end{equation}
This average is a Monte Carlo estimate of the expected bottom-level prediction under the hierarchy distribution and makes the final prediction less dependent on any single structure.
We describe how to design the hierarchy distribution $\mathcal P(\cdot\mid\mA)$ in~\autoref{method:distribution}, and the predictor $f_\theta$ in~\autoref{method:pyramidal}. Finally, in~\autoref{method:reconcile}, we consider the problem of enforcing aggregation constraints on the output of $f_\theta({}\cdot{})$ while also ensuring computational scalability.

\subsection{Designing the hierarchy distribution}
\label{method:distribution}
 
The distribution $\mathcal{P}(\cdot\mid\mA)$ controls the amount and nature of the structural variation introduced during training. Our framework does not prescribe a specific hierarchy distribution. A minimally constrained distribution may randomize both node assignments and the number of aggregates. Alternatively, hierarchy generation can preserve relationships that may be relevant to the task, such as data-driven similarities~\citep{pang_2022_hierarchical_electricity} or spatial dependencies encoded by a graph~\citep{cini_2024_higp,carotti_2026_graph_hierarchical_recurrence}. For the latter, structures can be generated through graph pooling methods \citep{grattarola_2022_understanding}. 
Among existing methods, \gls{kmis}~\citep{bacciu_2023_kmis_downsampling} generalizes downsampling to arbitrary irregular graphs by selecting centroids more than $k$ hops apart, while ensuring that every node lies within $k$ hops of a centroid.
As such, \gls{kmis} coarsening fits well within the framework as it preserves graph locality while allowing the aggregates to vary by considering different pooled graphs. 
In particular, different hierarchies can be obtained by changing $k$, which controls the number and spatial range of the intermediate aggregates, and the node ranking $\pi$, which changes the centroid selection.
In our implementation, we use three-level hierarchies in which \gls{kmis} maps bottom-level nodes into clusters at the second level. Then, all intermediate nodes are connected to a single top-level aggregate. Let $\mathcal K$ denote the candidate values of $k$ and let $\mathfrak S_N$ denote the set of permutations of the $N$ graph nodes. 
At each training step, we sample
$
k\sim\operatorname{Unif}(\mathcal K),
\pi\sim\operatorname{Unif}(\mathfrak S_N),
$
and construct the hierarchy
$
\mathcal H_{k,\pi}
=
\text{k-}\mathrm{MIS}(\mA,k,\pi).
$
This provides one possible design for the hierarchy distribution in~\autoref{eq:multi-hierarchy-objective}.
In our experiments, we primarily use this topology-aware distribution and also evaluate random node assignments (see \autoref{sec:experiments}).

\subsection{Pyramidal forecaster}
\label{method:pyramidal}

We instantiate the forecaster $f_\theta$ in \autoref{eq:multi-hierarchy-objective} using a variant of the pyramidal graph architecture of \citet{cini_2024_higp}. 
The architecture first encodes the history of every node in the hierarchy. 
At level $l$, a temporal encoder processes the corresponding time series and exogenous variables, producing representations $\mH_t^{(l),0}$. These are then processed by $B$ pyramidal layers. For simplicity, we omit the superscript $\mathcal H$ from quantities determined by the sampled hierarchy, including $\mH_t^{(l),b}$, $\mS^{(l)}$, and $\widehat{\mY}_{t+1:t+H}^{(l)}$. At the bottom level, each layer applies a \gls{gnn} to model dependencies among neighboring nodes, while information propagates between adjacent hierarchy levels through the assignment matrices
$\mS^{(l)}\in\{0,1\}^{N_{l-1}\times N_l}$. In particular, $\mS^{(l)\top}$ aggregates representations from level $l-1$ to level $l$, while $\mS^{(l)}$ propagates them in the opposite direction. Each layer $b\in[0,B-1]$ is defined as:
\begin{equation}
\begin{aligned}
    \mZ_t^{(l),b}
    &=
    \begin{cases}
        \mathrm{GNN}^{b}
        \left(\mH_t^{(0),b},\mA\right),
        & l=0,\\
        \mH_t^{(l),b},
        & l>0,
    \end{cases}
    \\
    \mH_t^{(l),b+1}
    &=
    \mathrm{Update}^{(l),b}
    \left(
        \mZ_t^{(l),b},
        \mS^{(l)\top}\mZ_t^{(l-1),b},
        \mS^{(l+1)}\mZ_t^{(l+1),b}
    \right).
\end{aligned}
\label{eq:pyramidal-message-passing}
\end{equation}
After $B$ layers, a \gls{mlp} maps each final node representation to its forecast.
The sampled hierarchy determines the aggregate representations $\mH_t^{(l),0}$ for $l>0$ and the assignment matrices $\mS^{(l)}$, thereby changing the aggregate nodes and propagation paths used by the model. 
We stack forecasts from all levels in $\widehat{\mY}_{t+1:t+H}$ and enforce coherence as described in~\autoref{method:reconcile}.

\subsection{Obtaining coherent forecasts efficiently}
\label{method:reconcile}
\begin{wrapfigure}{r}{0.5\linewidth}
\vspace{-1.8em}
  \centering
    \includegraphics[width=\linewidth]{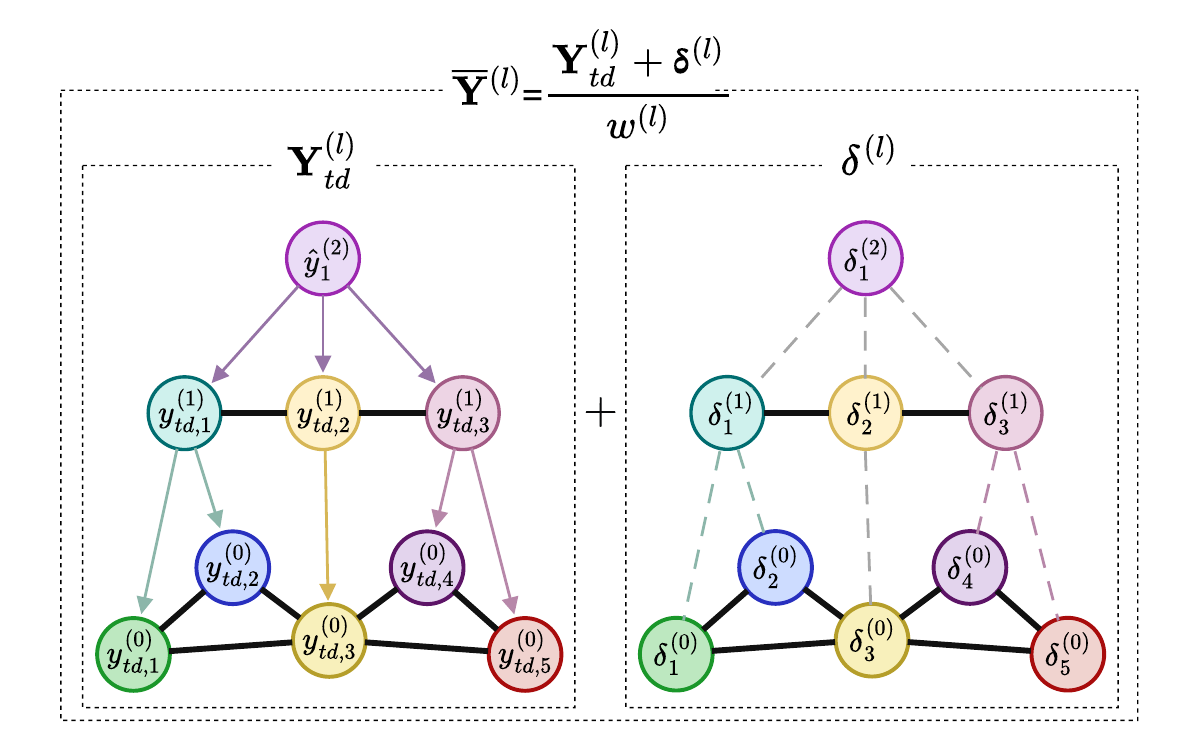}
  \vspace{-.5cm}
  \caption{Recursive operator for coherence.}
\vspace{-1em}
  \label{fig:new_riconciliation}
\end{wrapfigure}
The predictions produced by the pyramidal forecaster at every hierarchy node are not generally coherent across levels. For a hierarchy of $M$ nodes, the projection in~\autoref{eq:projection_reconciliation} requires a dense matrix inversion, with worst-case time and memory complexities of $\mathcal{O}(M^3)$ and $\mathcal{O}(M^2)$, respectively. When the hierarchy is fixed, the projection matrix can be computed once and reused throughout training~\citep{rangapuram_2021_endtoend}. When the hierarchy is learned, as in \citet{cini_2024_higp}, or resampled, as in our method, it must instead be recomputed whenever the structure changes. 
Therefore, computing coherent forecasts efficiently becomes particularly important in these settings. 
To this end, we impose coherence recursively between adjacent levels. At each step, a parent forecast is distributed among its children, and a zero-sum correction derived from the children’s forecasts is added without altering their aggregate.
For clarity, we omit the forecasting-horizon indices in the following.
Recall that $\mS^{(l)}\in\{0,1\}^{N_{l-1}\times N_l}$ assigns each node at level $l-1$ to one parent at level $l$.
Let $w_i^{(l)}$ be the number of bottom-level nodes represented by node $i$ at level $l$. Starting from $\vw^{(0)}=\mathbf{1}$, we compute $\vw^{(l+1)}=\mS^{(l+1)\top}\vw^{(l)}$ and rescale the forecasts according to their numbers of bottom-level descendants
$\widetilde{\mY}^{(l)}=\vw^{(l)}\odot\widehat{\mY}^{(l)}$.
We initialize $\overline{\mY}^{(L-1)}=\widetilde{\mY}^{(L-1)}$ and, for $l=L-2,\ldots,0$, recursively compute $\overline{\mY}^{(l)}=\mY_{\mathrm{td}}^{(l)}+\boldsymbol{\delta}^{(l)}$ as follows:
\begin{align}
    \mY_{\mathrm{td}}^{(l)}=
    \vw^{(l)}\odot\mS^{(l+1)}
    \left(
        \frac{\overline{\mY}^{(l+1)}}
             {\vw^{(l+1)}}
    \right),
    \quad
    \boldsymbol{\delta}^{(l)}=
    \widetilde{\mY}^{(l)}
    -
    \vw^{(l)}\odot\mS^{(l+1)}
    \left(
        \frac{\mS^{(l+1)\top}\widetilde{\mY}^{(l)}}
             {\vw^{(l+1)}}
    \right)
    \label{eq:top-down-rec}.
\end{align}
Finally, the resulting forecasts at each level are divided by the descendant counts to recover the original scale.
The top-down component $\mY_{\mathrm{td}}^{(l)}$ distributes each coherent parent forecast to its children proportionally to their numbers of bottom-level descendants. The correction $\boldsymbol{\delta}^{(l)}$ captures the weighted deviations of the lower-level forecasts from the average of their siblings. Since these corrections sum to zero within each sibling group, $\mS^{(l+1)\top}\boldsymbol{\delta}^{(l)}=\mathbf{0}$, they preserve coherence between adjacent levels (see \autoref{fig:new_riconciliation}). Repeating this update from the top level downward produces coherent forecasts at every level.
The procedure uses only local reductions from children to parents and broadcasts from parents to children, resulting in $\mathcal{O}(M)$ time and memory complexity.

\section{Related work}

\Glspl{stgnn} model temporal dynamics and spatial dependencies by combining temporal and graph-based operations~\citep{wu_2019_graph_wavenet,wu_2020_connecting,seo_2018_structured,yu_2021_stunet,LEI_2020_adaptive_gnn,yu_2018_spatio_temporal,li_2018_diffusion}.
Hierarchical structures have been used in \glspl{gnn} to pool nodes into coarser graphs progressively. This allows signals to be processed at multiple resolutions and facilitates long-range \gls{mp} through the coarser levels~\citep{ying_2018_hierarchical_graph,gao_2019_graph_unet,bianchi_2022_hierarchical_gnn,bianchi_2022_reservoir,carotti_2026_graph_hierarchical_recurrence}. This pyramidal design has also been adopted in spatiotemporal models \citep{yu_2021_stunet,hermes_2022_unet_traffic,guo_2021_hierarchical,ma_2022_hierarchical_pandemic,chen_2023_gnn_air_quality,arya_2023_hierarchical_stock,marisca_2024_graph_missing_downsampling}. 
A separate line of work forecasts series at multiple aggregation levels and enforces coherence among their predictions. \citet{rangapuram_2021_endtoend} incorporate predefined aggregation constraints into a probabilistic forecasting model, while \citet{cini_2024_higp} jointly learn the hierarchy with the forecasting model. Both approaches train the model using a single hierarchy.
Besides deep learning methods, recent work has begun to consider multiple aggregation structures. \citet{zhang_2025_hier_optimal_forecast} fit and reconcile forecasts separately under each hierarchy before combining the resulting predictions. However, applying this approach to deep models would require training a separate model for each hierarchy. Other methods build multiple aggregate groupings and use their constraints jointly during reconciliation~\citep{pang_2022_hierarchical_electricity,mattera_2024_improving}.
In both cases, hierarchy variation is used after forecasts have been produced to reconcile or combine predictions, rather than as a source of regularization during model training.

\section{Experiments}
\label{sec:experiments}

We aim to assess whether diversity in hierarchy structures can act as a source of regularization to improve forecasting accuracy, and whether our framework allows scalable training over a distribution of hierarchies. Specifically, we analyze: (1) whether training across multiple hierarchies improves accuracy over single-hierarchy alternatives and compares favorably with the baselines; (2) how the diversity of the hierarchies affects accuracy at training and inference time; (3)~whether our procedure to obtain coherence scales to large hierarchies; and (4) whether the derived regularization improves generalization to unseen series compared with fixed-hierarchy alternatives.
We refer to the framework introduced in \autoref{sec:method} as \gls{shr}. To isolate the effect of regularization from architectural choices, we instantiate the forecaster using the pyramidal architecture of \gls{higp}~\citep{cini_2024_higp}, with a \gls{rnn} for temporal encoding and gated graph convolutions \citep{cini_2023_taming} for \gls{mp}.
\Gls{shr} trains across sampled hierarchies (\autoref{eq:multi-hierarchy-objective}), enforces coherence with our recursive procedure (\autoref{method:reconcile}), and averages forecasts across hierarchies at inference (\autoref{eq:hierarchy_averaging}). In contrast, \gls{higp} is trained with a single hierarchy learned during training and uses the projection-based reconciliation (\autoref{eq:projection_reconciliation}). 
Unless otherwise specified, \gls{shr} uses the \gls{kmis}-based hierarchy distribution described in \autoref{method:distribution}; we also evaluate a random-assignment variant, denoted \textit{\gls{shr} (rand.)}.

\paragraph{Experimental settings}
We use four real-world sensor-network datasets with hundreds of nodes, covering traffic (\gls{la} and \gls{bay}, \cite{li_2018_diffusion}), air quality (\gls{air}, \cite{zheng_2015_forecasting}), and weather (\gls{engrad}, \cite{marisca_2024_graph_missing_downsampling}). We follow established settings from the literature for train, validation, and test splits, input windows, forecasting horizons, adjacency matrices, and exogenous variables \citep{cini_2023_taming,yi_2016_st,cini_2022_filling,marisca_2024_graph_missing_downsampling}.
For the inductive setting, we use the large-scale traffic dataset \gls{gla}, part of \gls{largest}~\citep{liu_2023_largest}.
Our baselines include \gls{tts} architectures with a GRU~\citep{cho_2014_gru} temporal encoder combined with different \gls{sota} \glspl{gnn}: bidirectional diffusion convolution~(\textbf{DCRNN}, \cite{li_2018_diffusion}), graph convolution~(\textbf{GConv-TTS}, \cite{kipf_2017_gconv}), and hierarchical Graph U-Net~(\textbf{GUNet-TTS}, \cite{gao_2019_graph_unet}). We also include a global univariate \gls{rnn} as a simple reference and several \gls{sota} \glspl{stgnn}: Graph WaveNet~(\textbf{GWNet}, \cite{wu_2019_graph_wavenet}), \textbf{SGP}~\citep{cini_2023_scalable}, \textbf{STAEformer}~\citep{liu_2023_staeformer},  \textbf{LAMP}~\citep{castelli_2026_adaptive}, \textbf{STDN}~\citep{cao_2025_stdn}, and \textbf{\gls{higp}}~\citep{cini_2024_higp}.
Full experimental details and additional results are provided in \autoref{a:data}, \autoref{a:baselines}, and \autoref{a:additional_results}.
For all experiments, we report the mean and standard deviation of the \gls{mae} across independent runs.

\begin{table*}[htbp]
\caption{Performance of baselines and \gls{shr} across datasets. \textbf{Bold} and \underline{underlined} denote the best and second-best mean \gls{mae}.}
\label{tab:baselines}

\centering
\resizebox{0.57\textwidth}{!}{%
\begin{tabular}{l|r|r|r|r}
\toprule
Model & \multicolumn{1}{c|}{\gls{la}} & \multicolumn{1}{c|}{\gls{bay}} & \multicolumn{1}{c|}{\gls{air}} & \multicolumn{1}{c}{\gls{engrad}} \\
\midrule
RNN & 3.540 {\tiny $\pm\,.001$} & 1.771 {\tiny $\pm\,.000$} & 13.919 {\tiny $\pm\,.052$} & 46.727 {\tiny $\pm\,.489$} \\
DCRNN & 3.281 {\tiny $\pm\,.001$} & 1.710 {\tiny $\pm\,.004$} & 12.685 {\tiny $\pm\,.018$} & 43.408 {\tiny $\pm\,.084$} \\
GConv-TTS & 3.071 {\tiny $\pm\,.004$} & 1.571 {\tiny $\pm\,.007$} & 12.393 {\tiny $\pm\,.020$} & 43.789 {\tiny $\pm\,.071$} \\
GUNet-TTS & 3.056 {\tiny $\pm\,.002$} & 1.571 {\tiny $\pm\,.003$} & 12.325 {\tiny $\pm\,.020$} & 44.179 {\tiny $\pm\,.439$} \\
GWNet & 3.069 {\tiny $\pm\,.003$} & 1.573 {\tiny $\pm\,.008$} & 11.962 {\tiny $\pm\,.033$} & 40.571 {\tiny $\pm\,.405$} \\
SGP & 3.047 {\tiny $\pm\,.002$} & 1.593 {\tiny $\pm\,.004$} & 12.727 {\tiny $\pm\,.010$} & 43.985 {\tiny $\pm\,.151$} \\
STAEformer & \underline{3.021 {\tiny $\pm\,.020$}} & \textbf{1.552 {\tiny $\pm\,.008$}} & 12.811 {\tiny $\pm\,.049$} & 57.418 {\tiny $\pm\,.801$} \\
LAMP & 3.132 {\tiny $\pm\,.001$} & 1.689 {\tiny $\pm\,.002$} & 15.827 {\tiny $\pm\,.028$} & 50.065 {\tiny $\pm\,.155$} \\
STDN & 3.050 {\tiny $\pm\,.007$} & 1.584 {\tiny $\pm\,.005$} & 13.222 {\tiny $\pm\,.133$} & 52.935 {\tiny $\pm\,.454$} \\
\gls{higp} & 3.035 {\tiny $\pm\,.008$} & 1.579 {\tiny $\pm\,.003$} & 11.984 {\tiny $\pm\,.017$} & 40.551 {\tiny $\pm\,.302$} \\
\midrule
\gls{shr} & \textbf{3.003 {\tiny $\pm\,.017$}} & \underline{1.556 {\tiny $\pm\,.008$}} & \textbf{11.945 {\tiny $\pm\,.004$}} & \textbf{39.167 {\tiny $\pm\,.160$}} \\
\bottomrule
\end{tabular}
}
\end{table*}
\begin{table*}[htbp]
\caption{Comparison of single- and multiple-hierarchy training, hierarchy construction strategies, and reconciliation procedures. \textbf{Bold} and \underline{underlined} denote the best and second-best mean \gls{mae}.}
\label{tab:baselines_ablation}
\centering
\resizebox{0.61\textwidth}{!}{
\begin{tabular}{l|r|r|r|r}
\toprule
Model & \multicolumn{1}{c|}{\gls{la}} & \multicolumn{1}{c|}{\gls{bay}} & \multicolumn{1}{c|}{\gls{air}} & \multicolumn{1}{c}{\gls{engrad}} \\
\midrule
Fixed hier. & 3.066 {\tiny $\pm\,.019$}  & 1.562 {\tiny $\pm\,.009$}  & 12.045 {\tiny $\pm\,.022$}  & 40.642 {\tiny $\pm\,.057$}  \\
HiGP & 3.035 {\tiny $\pm\,.008$}  & 1.579 {\tiny $\pm\,.003$}  & 11.984 {\tiny $\pm\,.017$}  & 40.551 {\tiny $\pm\,.302$}  \\
\midrule
\gls{shr} (proj.-based rec.) & 3.021 {\tiny $\pm\,.007$}  & 1.558 {\tiny $\pm\,.003$}  & \underline{11.951 {\tiny $\pm\,.049$}}  & \underline{40.002 {\tiny $\pm\,.180$}}  \\
\midrule
\gls{shr} (rand.) & \textbf{2.997 {\tiny $\pm\,.015$}} & \textbf{1.552 {\tiny $\pm\,.007$}} & 11.972 {\tiny $\pm\,.029$} & 40.170 {\tiny $\pm\,.256$} \\
\gls{shr} & \underline{3.003 {\tiny $\pm\,.017$}}  & \underline{1.556 {\tiny $\pm\,.008$}}  & \textbf{11.945 {\tiny $\pm\,.004$}}  & \textbf{39.167 {\tiny $\pm\,.160$}}  \\
\bottomrule
\end{tabular}
}
\end{table*}

\paragraph{Forecasting performance and hierarchy variants}
\autoref{tab:baselines} reports bottom-level forecasting performance for \gls{shr} and a range of established baselines and \gls{sota} models across datasets.
Despite its simple backbone, \gls{shr} is competitive with several established baselines.
To assess whether these benefits derive from training across multiple hierarchies, we compare \gls{shr} with two single-hierarchy variants using the same forecasting backbone in \autoref{tab:baselines_ablation}. \gls{higp} uses a hierarchy learnt jointly with the forecasting model, whereas \textit{Fixed hier.} uses a fixed hierarchy obtained through spectral clustering. \gls{shr} shows better performance than both, supporting the hypothesis that training across multiple aggregation structures can provide regularization for the bottom-level task.
In our experiments, we generally use \gls{kmis} to constrain node assignments through graph connectivity. We assess this choice with \textit{\gls{shr} (rand.)}, which randomly assigns bottom-level nodes to balanced aggregates. 
The fully randomized approach can perform well, although \gls{kmis} yields markedly better results in some settings, such as on \gls{engrad}.
The differences cannot be attributed solely to graph connectivity, as the two constructions also differ in the number, size, and balance of their aggregates. 
Finally, {\textit{\gls{shr}~(proj.-based rec.)}} replaces our recursive procedure in \gls{shr} with projection-based reconciliation (\autoref{eq:projection_reconciliation}), commonly adopted in deep hierarchical forecasting models~\citep{rangapuram_2021_endtoend,cini_2024_higp}. As shown in \autoref{tab:baselines_ablation}, our procedure achieves similar or better forecasting accuracy than standard projection-based reconciliation, showing that improved scalability can be achieved without compromising forecasting accuracy.

\begin{figure}[t]
    \centering
    \begin{subfigure}[t]{0.325\textwidth}
        \centering
        \includegraphics[width=\linewidth]{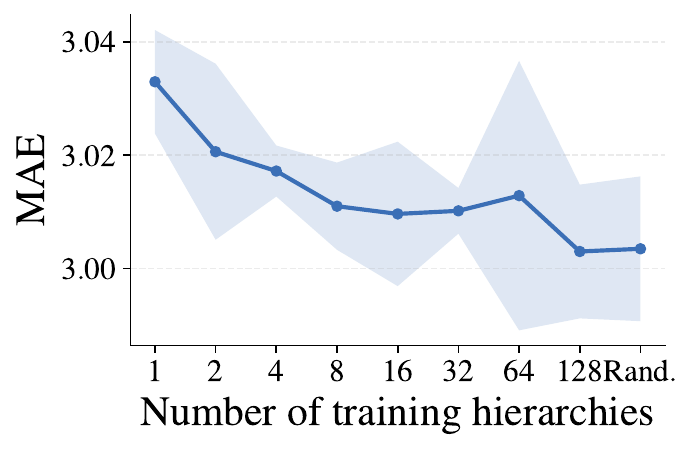}
        \caption{Fixed $N_1$, performance at $L_0$.}
        \label{fig:hierarchy_count_same_val_test_la_five_seeds}
    \end{subfigure}
    \hfill
    \begin{subfigure}[t]{0.325\textwidth}
        \centering
        \includegraphics[width=\linewidth]{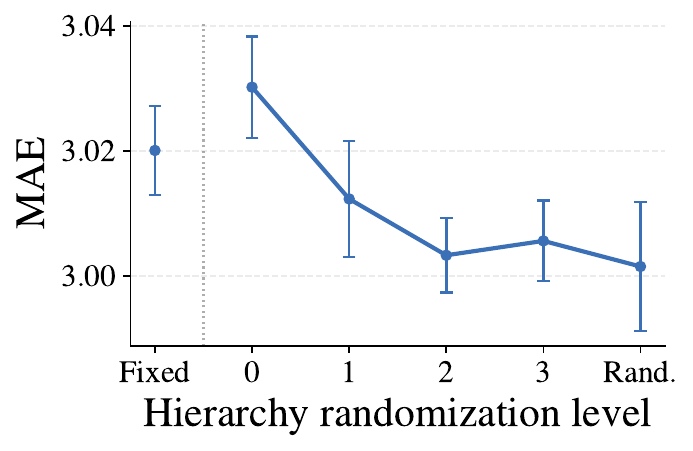}
        \caption{Variable $N_1$, performance at $L_0$.}
        \label{fig:progressive_same_val_test_la_five_seeds}
    \end{subfigure}
    \hfill
    \begin{subfigure}[t]{0.325\textwidth}
        \centering
        \includegraphics[width=\linewidth]{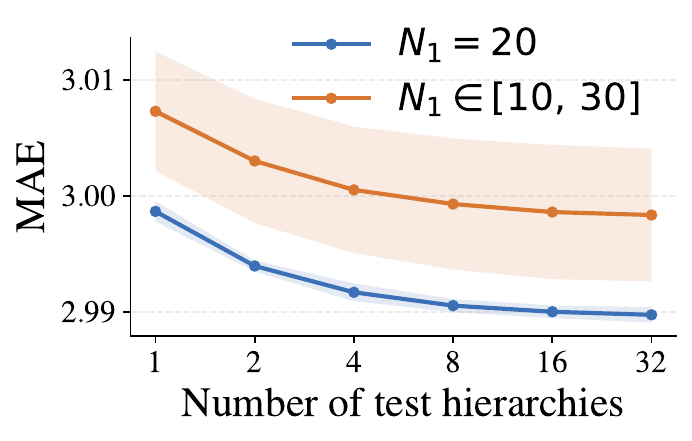}
        \caption{Test-time averaging at $L_0$.}
        \label{fig:inference_hierarchy_ensemble_la}
    \end{subfigure}
    \caption{Bottom-level performance ($L_0$) on \gls{la}. (a) varies the number of training hierarchies with fixed $N_1=20$; (b) increases the diversity of both node-to-aggregate assignments and the number of intermediate aggregates $N_1$; (c) varies the number of test-hierarchy forecasts being averaged, comparing fixed $N_1=20$ with $N_1\in\{10,\ldots,30\}$.
}
    \label{fig:hierarchy_variability}
\end{figure}

\begin{figure}[t]
    \centering
    \begin{subfigure}[t]{0.245\textwidth}
        \centering
        \includegraphics[width=\linewidth]{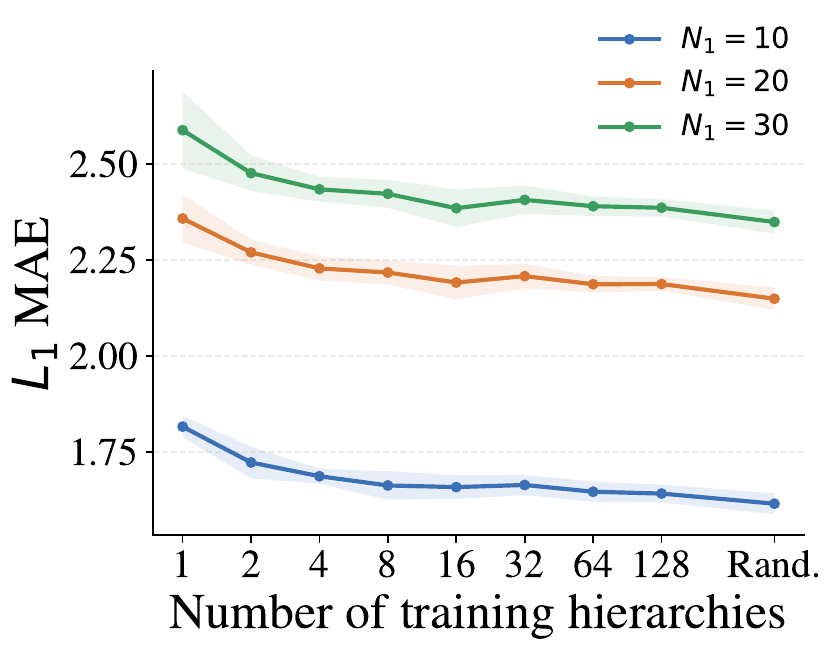}
        \caption{Fixed $N_1$ ($L_1$).}
        \label{fig:hierarchy_count_same_val_test_la_five_seeds_shadow_L1}
    \end{subfigure}
    \hfill
    \begin{subfigure}[t]{0.245\textwidth}
        \centering
        \includegraphics[width=\linewidth]{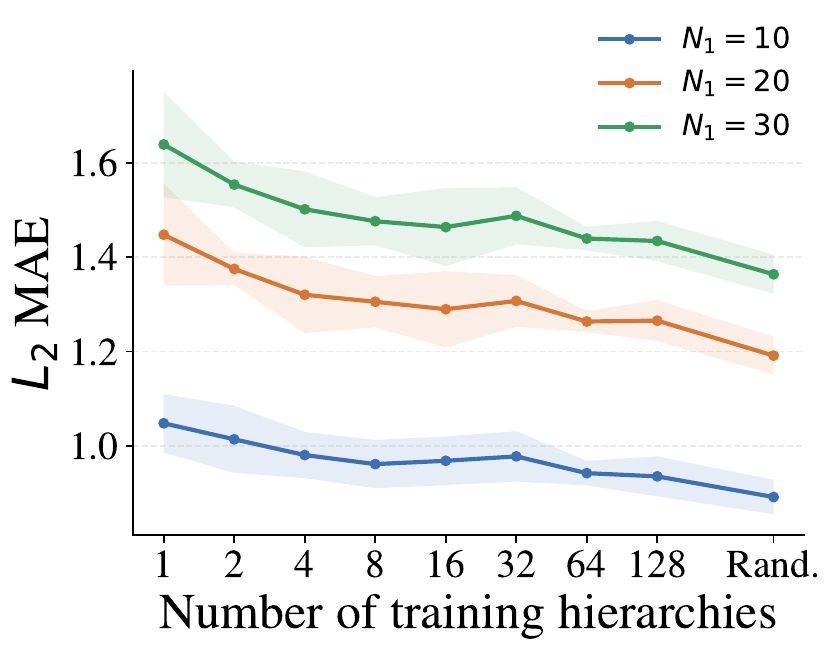}
        \caption{Fixed $N_1$ ($L_2$).}
        \label{fig:hierarchy_count_same_val_test_la_five_seeds_shadow_L2}
    \end{subfigure}
    \hfill
    \begin{subfigure}[t]{0.245\textwidth}
        \centering
        \includegraphics[width=\linewidth]{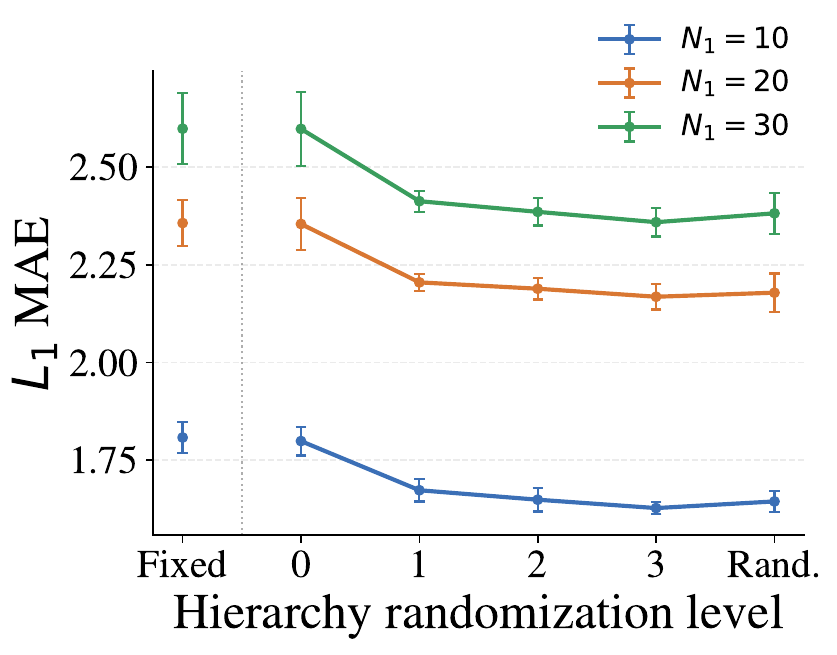}
        \caption{Variable $N_1$ ($L_1$).}
        \label{fig:progressive_same_val_test_la_five_seeds_L1}
    \end{subfigure}
    \hfill
    \begin{subfigure}[t]{0.245\textwidth}
        \centering
        \includegraphics[width=\linewidth]{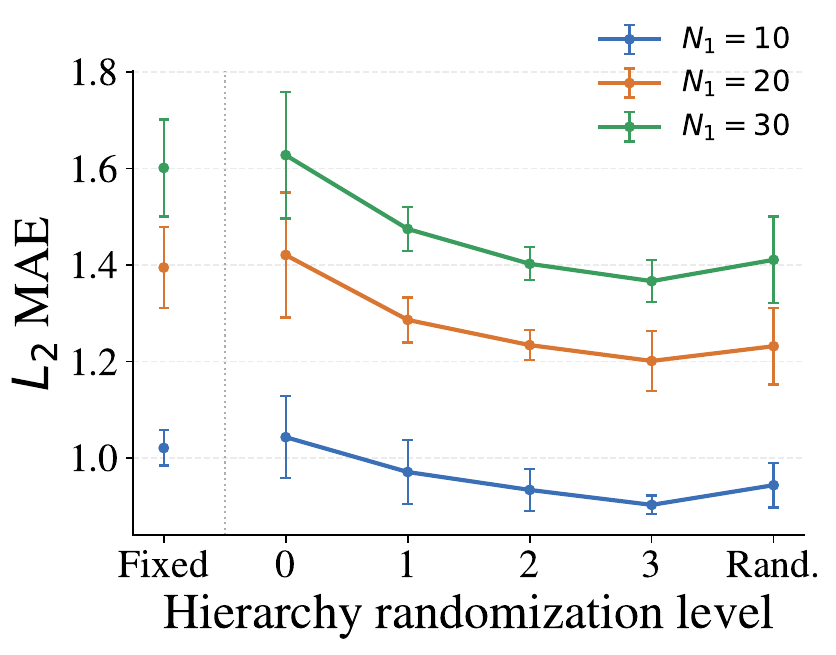}
        \caption{Variable $N_1$ ($L_2$).}
        \label{fig:progressive_same_val_test_la_five_seeds_L2}
    \end{subfigure}
    \caption{Performance at higher aggregation levels ($L_1$ and $L_2$) on \gls{la}. (a) and (b) increase the number of training hierarchies while keeping $N_1=20$ and varying node-to-aggregate assignments. (c) and (d) progressively increase training-hierarchy diversity by varying both the assignments and $N_1$. Each curve evaluates the same set of trained models using common test hierarchies with $N_1\in\{10,20,30\}$.
    }
    \label{fig:hierarchy_variability_higher_levels}
\end{figure}

\begin{table*}[htb]
\caption{Bottom-level performance under different hierarchy distributions induced by \gls{kmis}. \textbf{Bold} and \underline{underlined} denote the best and second-best mean \gls{mae}.}
\label{tab:ablation_variable_k}
\centering
\resizebox{0.7\textwidth}{!}{
\begin{tabular}{l|r|r|r|r}
\toprule
Model & \multicolumn{1}{c|}{\gls{la}} & \multicolumn{1}{c|}{\gls{bay}} & \multicolumn{1}{c|}{\gls{air}} & \multicolumn{1}{c}{\gls{engrad}} \\
\midrule
$k=2$, fixed ranking & 3.025 {\tiny $\pm\,.009$} & 1.559 {\tiny $\pm\,.013$} & 12.076 {\tiny $\pm\,.066$} & 39.769 {\tiny $\pm\,.142$} \\
$k=3$, fixed ranking & 3.020 {\tiny $\pm\,.013$} & 1.566 {\tiny $\pm\,.008$} & 12.054 {\tiny $\pm\,.012$} & 39.670 {\tiny $\pm\,.269$} \\
\midrule
$k=2$, random ranking & 3.027 {\tiny $\pm\,.031$} & \textbf{1.552 {\tiny $\pm\,.003$}} & \textbf{11.940 {\tiny $\pm\,.012$}} & \underline{39.317 {\tiny $\pm\,.101$}} \\
$k=3$, random ranking & \textbf{3.012 {\tiny $\pm\,.010$}} & 1.559 {\tiny $\pm\,.001$} & 11.987 {\tiny $\pm\,.027$} & \textbf{39.233 {\tiny $\pm\,.361$}} \\
$k\in\{2,3\}$, random ranking & \underline{3.014 {\tiny $\pm\,.035$}} & \underline{1.553 {\tiny $\pm\,.011$}} & \underline{11.979 {\tiny $\pm\,.021$}} & 39.358 {\tiny $\pm\,.196$} \\
\bottomrule
\end{tabular}
}
\end{table*}

\paragraph{Sensitivity analysis.}
As described in \autoref{method:inductive}, \gls{shr} leverages hierarchy diversity during both training and inference. During training, a single model learns across different aggregation structures, while at inference its bottom-level forecasts can be averaged across hierarchies. 
Here, we investigate in depth whether increasing hierarchy diversity during training benefits forecasting accuracy. We use the \gls{la} dataset as reference and provide results for other datasets in \autoref{a:additional_results}. In \autoref{fig:hierarchy_count_same_val_test_la_five_seeds}, we progressively increase the number of training hierarchies while keeping the number of intermediate nodes fixed ($N_1=20$) and varying their bottom-level assignments. One hierarchy is sampled at each training batch from a finite set of increasing size, while \textit{Rand.} generates a new assignment every time. 
Increasing the number of training hierarchies generally reduces \gls{mae}, supporting the interpretation that different structures may provide useful learning signals for the downstream task.
We then study whether introducing broader structural diversity provides similar benefits. In \autoref{fig:progressive_same_val_test_la_five_seeds}, we progressively increase hierarchy diversity by varying both bottom-level assignments and the number of intermediate nodes.
\textit{Hierarchy randomization Level}~0 and~1 use one and four hierarchies with $N_1=20$. Levels~2 and~3 use 16 and 64 hierarchies with $N_1\in\{15,20,25\}$ and $N_1\in\{10,15,20,25,30\}$, respectively. At Level~\textit{Rand.} (corresponding to \textit{\gls{shr} (rand.)} in \autoref{tab:baselines_ablation}), a new hierarchy is generated at every training batch with $N_1\in\{10,\ldots,30\}$.
In both experiments, the compared models use three randomly generated hierarchies per validation batch and report bottom-level \gls{mae} after averaging forecasts from multiple hierarchies. 
\textit{Fixed} instead uses a single hierarchy for training, validation, and testing. Also in this case, training on multiple, structurally different hierarchies improves bottom-level accuracy. This trend persists without test-time averaging (\autoref{a:additional_results}). Similar gains at higher aggregation levels under three different hierarchies show that the benefit extends to aggregate forecasts under different aggregation structures~(\autoref{fig:hierarchy_variability_higher_levels}).
We next assess whether hierarchy diversity also provides benefits at inference through forecast averaging. 
In \autoref{fig:inference_hierarchy_ensemble_la}, we progressively increase the number of test hierarchies whose bottom-level forecasts are averaged according to \autoref{eq:hierarchy_averaging}, using two separately trained \textit{\gls{shr} (rand.)} models. For the first, every training and test hierarchy has $N_1=20$; for the second, $N_1$ is independently sampled from $\{10,\ldots,30\}$ for each training batch and test hierarchy.
Accuracy consistently improves as more forecasts are averaged, both for fixed and varying $N_1$. 
Finally, we investigate whether training across multiple hierarchies is also beneficial when the hierarchy distribution is constrained by graph topology.
\autoref{tab:ablation_variable_k} compares different hierarchy distributions induced by \gls{kmis}.
Fixed $k$ and node rankings $\pi$ define a single hierarchy, whereas randomized rankings generate multiple hierarchies by changing centroid selection and node assignments. We consider fixed and randomized rankings for $k\in\{2,3\}$, as well as randomized rankings with sampled $k$. Randomized rankings perform best on most datasets, particularly \gls{air} and \gls{engrad}, showing that the benefits of training across multiple hierarchies persist across different hierarchy distributions.
\begin{figure}[t]
    \centering
    \vspace{-0.5em}
    \begin{subfigure}[t]{0.4\textwidth}
        \centering
        \includegraphics[width=\linewidth]{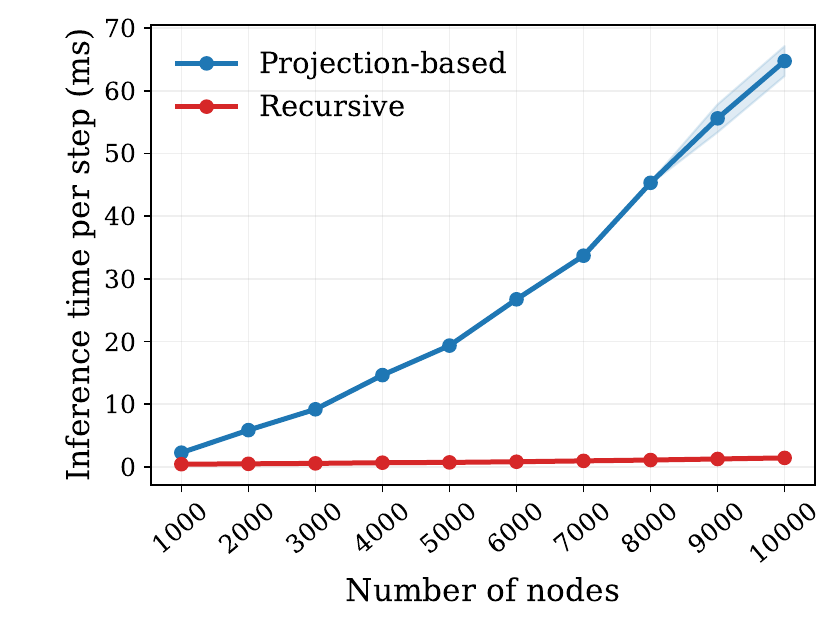}
    \end{subfigure}
    \begin{subfigure}[t]{0.4\textwidth}
        \centering
        \includegraphics[width=\linewidth]{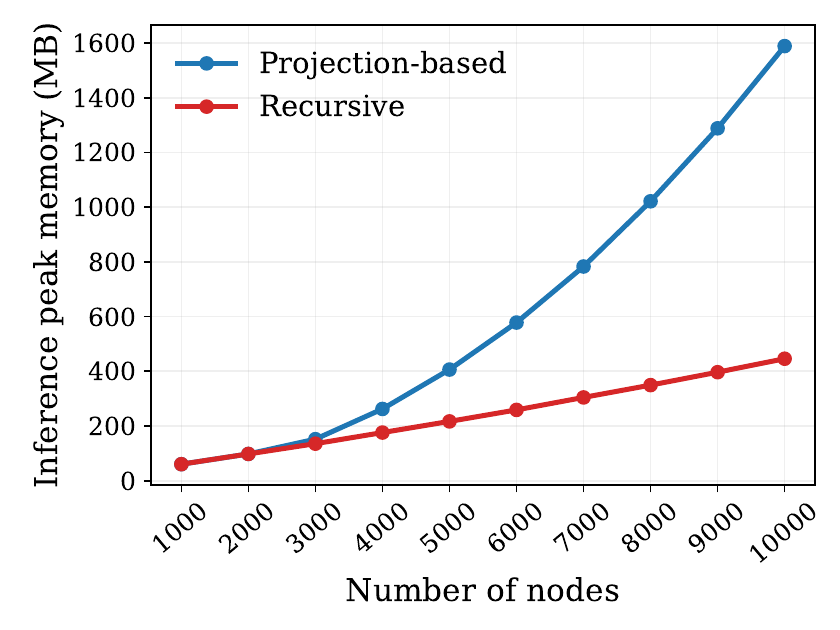}
    \end{subfigure}
    \caption{Inference scalability with projection-based reconciliation and our recursive operator, measured by runtime (left) and peak GPU memory (right) as graph and hierarchy sizes increase.}
    \label{fig:inference_scaling_cuda_time_peak_memory}
\end{figure}
\paragraph{Computational scalability.}
We assess how our recursive operator for obtaining coherent forecasts scales compared with the projection-based reconciliation in \autoref{eq:projection_reconciliation}, commonly used in deep hierarchical forecasting models~\citep{rangapuram_2021_endtoend,cini_2024_higp}. We use identical forecasting models that differ only in how coherence is enforced, and evaluate them on synthetic graphs and hierarchies of increasing size. As shown in \autoref{fig:inference_scaling_cuda_time_peak_memory}, our procedure requires substantially less inference time and GPU memory, with the gap widening at larger scales. Consistent with \autoref{method:reconcile}, its time and memory complexities grow approximately linearly, whereas projection-based reconciliation grows faster. Together with \autoref{tab:baselines_ablation}, these findings show that our procedure improves scalability relative to a dense projection implementation without sacrificing predictive performance. 
\paragraph{Inductive setting.}
\begin{wraptable}{r}{0.35\textwidth}
\centering
\vspace{-1.em}
\caption{Performance on \gls{gla}~(inductive setup). \textbf{Bold} and \underline{underlined} values indicate the best and second-best \gls{mae}. }
\label{tab:largest_inductive_short}
\vspace{-0.3em}
\resizebox{\linewidth}{!}{
\begin{tabular}{l|r|r}
\toprule
Model & Window & MAE \\
\midrule
Fixed hier. & 96 & 22.444 {\tiny $\pm\,.198$} \\
\midrule
\gls{shr} (rand.) & 96 & \textbf{21.240 {\tiny $\pm\,.057$}} \\
\gls{shr} & 96 & \underline{21.412 {\tiny $\pm\,.087$}} \\
\bottomrule
\end{tabular}
}
\vspace{-1.em}
\end{wraptable}
We evaluate \gls{shr} in an inductive setting by partitioning \gls{gla} into disjoint subgraphs for training, validation, and testing (further details in \autoref{a:data}). We compare \gls{shr} with a single-hierarchy model.
We exclude \gls{higp} from the comparison because it relies on node embeddings to construct the hierarchy \citep{cini_2023_taming}, which cannot directly transfer to unseen graphs. As shown in \autoref{tab:largest_inductive_short}, both \gls{shr} variants outperform the single-hierarchy model. 
Results suggest that \gls{shr} can support settings where the underlying graph and associated aggregation structure change.

\section{Conclusion}
We propose structurally diverse hierarchies as a source of regularization for spatiotemporal forecasting. We introduce a training framework in which a single model learns across multiple aggregation structures, together with a scalable procedure for ensuring coherent forecasts under each hierarchy. Experiments show consistent improvements over single-hierarchy training and competitive performance against forecasting baselines.
Future work could explore different distributions over hierarchies and different designs for the forecasting architecture. It would also be interesting to consider different aggregation levels jointly across time and space, incorporating temporal hierarchies~\citep{athanasopoulos_2017_forecasting,rangapuram_2023_coherent} into the framework.

\section*{Acknowledgments}
This work was supported by the Swiss National Science Foundation projects no.\ 204061~(Higher-Order Relations and Dynamics in Graph Neural Networks) and no.\ 225351~(Relational Deep Learning for Reliable Time Series Forecasting at Scale).

\bibliography{main}

@article{mattera_2024_improving,
author = {Raffaele Mattera and George Athanasopoulos and Rob Hyndman},
title = {Improving out-of-sample forecasts of stock price indexes with forecast reconciliation and clustering},
journal = {Quantitative Finance},
volume = {24},
number = {11},
pages = {1641--1667},
year = {2024},
publisher = {Routledge},
doi = {10.1080/14697688.2024.2412687},
URL = {https://doi.org/10.1080/14697688.2024.2412687},
eprint = {https://doi.org/10.1080/14697688.2024.2412687}}

@article{hollyman_2021_understanding_reconciliation,
title = {Understanding forecast reconciliation},
journal = {European Journal of Operational Research},
volume = {294},
number = {1},
pages = {149-160},
year = {2021},
issn = {0377-2217},
doi = {https://doi.org/10.1016/j.ejor.2021.01.017},
url = {https://www.sciencedirect.com/science/article/pii/S0377221721000199},
author = {Ross Hollyman and Fotios Petropoulos and Michael E. Tipping}
}

@article{bates_1969_forecast_combination,
 ISSN = {14732858},
 URL = {http://www.jstor.org/stable/3008764},
 author = {J. M. Bates and C. W. J. Granger},
 journal = {OR},
 number = {4},
 pages = {451--468},
 publisher = {Operational Research Society},
 title = {The Combination of Forecasts},
 urldate = {2026-07-30},
 volume = {20},
 year = {1969}
}

@article{zhang_2025_hier_optimal_forecast,
title = {Constructing hierarchical time series through clustering: Is there an optimal way for forecasting?},
journal = {International Journal of Forecasting},
volume = {41},
number = {3},
pages = {1022-1036},
year = {2025},
issn = {0169-2070},
doi = {https://doi.org/10.1016/j.ijforecast.2024.10.002},
url = {https://www.sciencedirect.com/science/article/pii/S0169207024001031},
author = {Bohan Zhang and Anastasios Panagiotelis and Han Li}
}

@inproceedings{cini_2024_higp,
author = {Cini, Andrea and Mandic, Danilo and Alippi, Cesare},
title = {Graph-based time series clustering for end-to-end hierarchical forecasting},
year = {2024},
publisher = {JMLR.org},
booktitle = {Proceedings of the 41st International Conference on Machine Learning},
articleno = {359},
numpages = {15},
location = {Vienna, Austria},
series = {ICML'24}
}

@article{hyndman_2016_fast_reconciled,
title = {Fast computation of reconciled forecasts for hierarchical and grouped time series},
journal = {Computational Statistics \& Data Analysis},
volume = {97},
pages = {16-32},
year = {2016},
issn = {0167-9473},
doi = {https://doi.org/10.1016/j.csda.2015.11.007},
url = {https://www.sciencedirect.com/science/article/pii/S016794731500290X},
author = {Rob J. Hyndman and Alan J. Lee and Earo Wang}
}

@article{sprangers_2024_forecasting_at_scale,
title = {Hierarchical forecasting at scale},
journal = {International Journal of Forecasting},
volume = {40},
number = {4},
pages = {1689-1700},
year = {2024},
issn = {0169-2070},
doi = {https://doi.org/10.1016/j.ijforecast.2024.02.006},
url = {https://www.sciencedirect.com/science/article/pii/S0169207024000116},
author = {Olivier Sprangers and Wander Wadman and Sebastian Schelter and Maarten {de Rijke}}
}

@article{petropoulos_2025_scalable_sub_hierarchies,
author = {Fotios Petropoulos and Ross A Hollyman and Evangelos Spiliotis},
title = {Scalable forecast reconciliation through (un)guided Sub-hierarchies},
journal = {Journal of the Operational Research Society},
volume = {0},
number = {0},
pages = {1--20},
year = {2025},
publisher = {Taylor \& Francis},
doi = {10.1080/01605682.2025.2599394},
URL = { 
        https://doi.org/10.1080/01605682.2025.2599394
},
eprint = {
        https://doi.org/10.1080/01605682.2025.2599394
}

}

@article{sharma_2025_hierarchical_network_flow,
  title={Hierarchical Forecast Reconciliation on Networks: A Network Flow Optimization Formulation},
  author={Sharma, Charupriya and Aguerri, I{\~n}aki Estella and Guimarans, Daniel},
  journal={arXiv preprint arXiv:2505.03955},
  year={2025}
}

@InProceedings{rangapuram_2021_endtoend,
  title = 	 {End-to-End Learning of Coherent Probabilistic Forecasts for Hierarchical Time Series},
  author =       {Rangapuram, Syama Sundar and Werner, Lucien D and Benidis, Konstantinos and Mercado, Pedro and Gasthaus, Jan and Januschowski, Tim},
  booktitle = 	 {Proceedings of the 38th International Conference on Machine Learning},
  pages = 	 {8832--8843},
  year = 	 {2021},
  editor = 	 {Meila, Marina and Zhang, Tong},
  volume = 	 {139},
  series = 	 {Proceedings of Machine Learning Research},
  month = 	 {18--24 Jul},
  publisher =    {PMLR},
  url = 	 {https://proceedings.mlr.press/v139/rangapuram21a.html}
}

@article{cini_2025_graph_deep,
author = {Cini, Andrea and Marisca, Ivan and Zambon, Daniele and Alippi, Cesare},
title = {Graph Deep Learning for Time Series Forecasting},
year = {2025},
issue_date = {December 2025},
publisher = {Association for Computing Machinery},
address = {New York, NY, USA},
volume = {57},
number = {12},
issn = {0360-0300},
url = {https://doi.org/10.1145/3742784},
doi = {10.1145/3742784},
journal = {ACM Comput. Surv.},
month = jul,
articleno = {321},
numpages = {34}
}

@inproceedings{caruana_1993_multitask,
author = {Caruana, Rich},
title = {Multitask learning: a knowledge-based source of inductive bias},
year = {1993},
isbn = {1558603077},
publisher = {Morgan Kaufmann Publishers Inc.},
address = {San Francisco, CA, USA},
booktitle = {Proceedings of the Tenth International Conference on Machine Learning},
pages = {41–48},
numpages = {8},
location = {Amherst, MA, USA},
series = {ICML'93}
}

@inproceedings{cini_2023_taming,
 author = {Cini, Andrea and Marisca, Ivan and Zambon, Daniele and Alippi, Cesare},
 booktitle = {Advances in Neural Information Processing Systems},
 doi = {10.52202/075280-2417},
 editor = {A. Oh and T. Naumann and A. Globerson and K. Saenko and M. Hardt and S. Levine},
 pages = {55375--55393},
 publisher = {Curran Associates, Inc.},
 title = {Taming Local Effects in Graph-based Spatiotemporal Forecasting},
 url = {https://proceedings.neurips.cc/paper_files/paper/2023/file/ad58c61c71efd5436134a3ecc87da6ea-Paper-Conference.pdf},
 volume = {36},
 year = {2023}
}

@InProceedings{gao_2021_equivalence,
  title = 	 {On the Equivalence Between Temporal and Static Equivariant Graph Representations},
  author =       {Gao, Jianfei and Ribeiro, Bruno},
  booktitle = 	 {Proceedings of the 39th International Conference on Machine Learning},
  pages = 	 {7052--7076},
  year = 	 {2022},
  editor = 	 {Chaudhuri, Kamalika and Jegelka, Stefanie and Song, Le and Szepesvari, Csaba and Niu, Gang and Sabato, Sivan},
  volume = 	 {162},
  series = 	 {Proceedings of Machine Learning Research},
  month = 	 {17--23 Jul},
  publisher =    {PMLR},
  url = 	 {https://proceedings.mlr.press/v162/gao22e.html}
}

@article{panagiotelis_2021_reconciliation_geometric,
title = {Forecast reconciliation: A geometric view with new insights on bias correction},
journal = {International Journal of Forecasting},
volume = {37},
number = {1},
pages = {343-359},
year = {2021},
issn = {0169-2070},
doi = {https://doi.org/10.1016/j.ijforecast.2020.06.004},
url = {https://www.sciencedirect.com/science/article/pii/S0169207020300911},
author = {Anastasios Panagiotelis and George Athanasopoulos and Puwasala Gamakumara and Rob J. Hyndman}
}

@article{panagiotelis_2023_probabilistic_reconciliation,
title = {Probabilistic forecast reconciliation: Properties, evaluation and score optimisation},
journal = {European Journal of Operational Research},
volume = {306},
number = {2},
pages = {693-706},
year = {2023},
issn = {0377-2217},
doi = {https://doi.org/10.1016/j.ejor.2022.07.040},
url = {https://www.sciencedirect.com/science/article/pii/S0377221722006087},
author = {Anastasios Panagiotelis and Puwasala Gamakumara and George Athanasopoulos and Rob J. Hyndman}
}

@article{wickramasuriya_2019_trace_minimization,
author = {Shanika L. Wickramasuriya and George Athanasopoulos and Rob J. Hyndman},
title = {Optimal Forecast Reconciliation for Hierarchical and Grouped Time Series Through Trace Minimization},
journal = {Journal of the American Statistical Association},
volume = {114},
number = {526},
pages = {804--819},
year = {2019},
publisher = {Taylor \& Francis},
doi = {10.1080/01621459.2018.1448825},


URL = { 
    
        https://doi.org/10.1080/01621459.2018.1448825
    
    

},
eprint = { 
    
        https://doi.org/10.1080/01621459.2018.1448825
    
    

}

}

@article{hyndman_2011_optimal,
title = {Optimal combination forecasts for hierarchical time series},
journal = {Computational Statistics \& Data Analysis},
volume = {55},
number = {9},
pages = {2579-2589},
year = {2011},
issn = {0167-9473},
doi = {https://doi.org/10.1016/j.csda.2011.03.006},
url = {https://www.sciencedirect.com/science/article/pii/S0167947311000971},
author = {Rob J. Hyndman and Roman A. Ahmed and George Athanasopoulos and Han Lin Shang}
}

@ARTICLE{grattarola_2022_understanding,
  author={Grattarola, Daniele and Zambon, Daniele and Bianchi, Filippo Maria and Alippi, Cesare},
  journal={IEEE Transactions on Neural Networks and Learning Systems}, 
  title={Understanding Pooling in Graph Neural Networks}, 
  year={2024},
  volume={35},
  number={2},
  pages={2708-2718},
  doi={10.1109/TNNLS.2022.3190922}}

@inproceedings{bacciu_2023_kmis_downsampling,
author = {Bacciu, Davide and Conte, Alessio and Landolfi, Francesco},
title = {Generalizing downsampling from regular data to graphs},
year = {2023},
isbn = {978-1-57735-880-0},
publisher = {AAAI Press},
url = {https://doi.org/10.1609/aaai.v37i6.25824},
doi = {10.1609/aaai.v37i6.25824},
booktitle = {Proceedings of the Thirty-Seventh AAAI Conference on Artificial Intelligence and Thirty-Fifth Conference on Innovative Applications of Artificial Intelligence and Thirteenth Symposium on Educational Advances in Artificial Intelligence},
articleno = {755},
numpages = {10},
series = {AAAI'23/IAAI'23/EAAI'23}
}

@article{yang_2017_reconciling,
title = {Reconciling solar forecasts: Geographical hierarchy},
journal = {Solar Energy},
volume = {146},
pages = {276-286},
year = {2017},
issn = {0038-092X},
doi = {https://doi.org/10.1016/j.solener.2017.02.010},
url = {https://www.sciencedirect.com/science/article/pii/S0038092X17301020},
author = {Dazhi Yang and Hao Quan and Vahid R. Disfani and Licheng Liu}
}

@article{pang_2022_hierarchical_electricity,
title = {Hierarchical electricity time series prediction with cluster analysis and sparse penalty},
journal = {Pattern Recognition},
volume = {126},
pages = {108555},
year = {2022},
issn = {0031-3203},
doi = {https://doi.org/10.1016/j.patcog.2022.108555},
url = {https://www.sciencedirect.com/science/article/pii/S003132032200036X},
author = {Yue Pang and Xiangdong Zhou and Junqi Zhang and Quan Sun and Jianbin Zheng}
}

@misc{carotti_2026_graph_hierarchical_recurrence,
      title={Graph Hierarchical Recurrence for Long-Range Generalization}, 
      author={Stefano Carotti and Marco Pacini and Alessio Gravina and Davide Bacciu and Bruno Lepri and Sebastiano Bontorin},
      year={2026},
      eprint={2605.18387},
      archivePrefix={arXiv},
      primaryClass={cs.LG},
      url={https://arxiv.org/abs/2605.18387}, 
}

@article{difonzo_2024_combination_reconciliation,
title = {Forecast combination-based forecast reconciliation: Insights and extensions},
journal = {International Journal of Forecasting},
volume = {40},
number = {2},
pages = {490-514},
year = {2024},
issn = {0169-2070},
doi = {https://doi.org/10.1016/j.ijforecast.2022.07.001},
url = {https://www.sciencedirect.com/science/article/pii/S0169207022000991},
author = {Tommaso {Di Fonzo} and Daniele Girolimetto}
}

@ARTICLE{jin_2024_survey_gnn,
  author={Jin, Ming and Koh, Huan Yee and Wen, Qingsong and Zambon, Daniele and Alippi, Cesare and Webb, Geoffrey I. and King, Irwin and Pan, Shirui},
  journal={IEEE Transactions on Pattern Analysis and Machine Intelligence}, 
  title={A Survey on Graph Neural Networks for Time Series: Forecasting, Classification, Imputation, and Anomaly Detection}, 
  year={2024},
  volume={46},
  number={12},
  pages={10466-10485},
  doi={10.1109/TPAMI.2024.3443141}}

@inproceedings{LEI_2020_adaptive_gnn,
 author = {BAI, LEI and Yao, Lina and Li, Can and Wang, Xianzhi and Wang, Can},
 booktitle = {Advances in Neural Information Processing Systems},
 editor = {H. Larochelle and M. Ranzato and R. Hadsell and M.F. Balcan and H. Lin},
 pages = {17804--17815},
 publisher = {Curran Associates, Inc.},
 title = {Adaptive Graph Convolutional Recurrent Network for Traffic Forecasting},
 url = {https://proceedings.neurips.cc/paper_files/paper/2020/file/ce1aad92b939420fc17005e5461e6f48-Paper.pdf},
 volume = {33},
 year = {2020}
}

@inproceedings{yu_2018_spatio_temporal,
author = {Yu, Bing and Yin, Haoteng and Zhu, Zhanxing},
title = {Spatio-temporal graph convolutional networks: a deep learning framework for traffic forecasting},
year = {2018},
isbn = {9780999241127},
publisher = {AAAI Press},
booktitle = {Proceedings of the 27th International Joint Conference on Artificial Intelligence},
pages = {3634–3640},
numpages = {7},
location = {Stockholm, Sweden},
series = {IJCAI'18}
}

@inproceedings{wu_2019_graph_wavenet,
author = {Wu, Zonghan and Pan, Shirui and Long, Guodong and Jiang, Jing and Zhang, Chengqi},
title = {Graph wavenet for deep spatial-temporal graph modeling},
year = {2019},
isbn = {9780999241141},
publisher = {AAAI Press},
booktitle = {Proceedings of the 28th International Joint Conference on Artificial Intelligence},
pages = {1907–1913},
numpages = {7},
location = {Macao, China},
series = {IJCAI'19}
}

@inproceedings{wu_2020_connecting,
author = {Wu, Zonghan and Pan, Shirui and Long, Guodong and Jiang, Jing and Chang, Xiaojun and Zhang, Chengqi},
title = {Connecting the Dots: Multivariate Time Series Forecasting with Graph Neural Networks},
year = {2020},
isbn = {9781450379984},
publisher = {Association for Computing Machinery},
address = {New York, NY, USA},
url = {https://doi.org/10.1145/3394486.3403118},
doi = {10.1145/3394486.3403118},
booktitle = {Proceedings of the 26th ACM SIGKDD International Conference on Knowledge Discovery \& Data Mining},
pages = {753–763},
numpages = {11},
location = {Virtual Event, CA, USA},
series = {KDD '20}
}

@InProceedings{seo_2018_structured,
author={Seo, Youngjoo
and Defferrard, Micha{\"e}l
and Vandergheynst, Pierre
and Bresson, Xavier},
editor={Cheng, Long
and Leung, Andrew Chi Sing
and Ozawa, Seiichi},
title={Structured Sequence Modeling with Graph Convolutional Recurrent Networks},
booktitle={Neural Information Processing},
year={2018},
publisher={Springer International Publishing},
address={Cham},
pages={362--373},
isbn={978-3-030-04167-0}
}

@misc{yu_2021_stunet,
      title={ST-UNet: A Spatio-Temporal U-Network for Graph-structured Time Series Modeling}, 
      author={Bing Yu and Haoteng Yin and Zhanxing Zhu},
      year={2021},
      eprint={1903.05631},
      archivePrefix={arXiv},
      primaryClass={cs.LG},
      url={https://arxiv.org/abs/1903.05631}, 
}

@inproceedings{marisca_2024_graph_missing_downsampling,
author = {Marisca, Ivan and Alippi, Cesare and Bianchi, Filippo Maria},
title = {Graph-based forecasting with missing data through spatiotemporal downsampling},
year = {2024},
publisher = {JMLR.org},
booktitle = {Proceedings of the 41st International Conference on Machine Learning},
articleno = {1417},
numpages = {20},
location = {Vienna, Austria},
series = {ICML'24}
}

@inproceedings{ma_2022_hierarchical_pandemic,
author = {Ma, Yihong and Gerard, Patrick and Tian, Yijun and Guo, Zhichun and Chawla, Nitesh V.},
title = {Hierarchical Spatio-Temporal Graph Neural Networks for Pandemic Forecasting},
year = {2022},
isbn = {9781450392365},
publisher = {Association for Computing Machinery},
address = {New York, NY, USA},
url = {https://doi.org/10.1145/3511808.3557350},
doi = {10.1145/3511808.3557350},
booktitle = {Proceedings of the 31st ACM International Conference on Information \& Knowledge Management},
pages = {1481–1490},
numpages = {10},
location = {Atlanta, GA, USA},
series = {CIKM '22}
}

@inproceedings{guo_2021_hierarchical,
  title={Hierarchical graph convolution network for traffic forecasting},
  author={Guo, Kan and Hu, Yongli and Sun, Yanfeng and Qian, Sean and Gao, Junbin and Yin, Baocai},
  booktitle={Proceedings of the AAAI conference on artificial intelligence},
  volume={35},
  pages={151--159},
  year={2021}
}

@inproceedings{cho_2014_gru,
    title = "On the Properties of Neural Machine Translation: Encoder{--}Decoder Approaches",
    author = {Cho, Kyunghyun  and
      van Merri{\"e}nboer, Bart  and
      Bahdanau, Dzmitry  and
      Bengio, Yoshua},
    editor = "Wu, Dekai  and
      Carpuat, Marine  and
      Carreras, Xavier  and
      Vecchi, Eva Maria",
    booktitle = "Proceedings of {SSST}-8, Eighth Workshop on Syntax, Semantics and Structure in Statistical Translation",
    month = oct,
    year = "2014",
    address = "Doha, Qatar",
    publisher = "Association for Computational Linguistics",
    url = "https://aclanthology.org/W14-4012/",
    doi = "10.3115/v1/W14-4012",
    pages = "103--111"
}

@INPROCEEDINGS{hermes_2022_unet_traffic,
  author={Hermes, Luca and Hammer, Barbara and Melnik, Andrew and Velioglu, Riza and Vieth, Markus and Schilling, Malte},
  booktitle={2022 International Joint Conference on Neural Networks (IJCNN)}, 
  title={A Graph-based U-Net Model for Predicting Traffic in unseen Cities}, 
  year={2022},
  volume={},
  number={},
  pages={1-8},
  doi={10.1109/IJCNN55064.2022.9892453}}

@article{chen_2023_gnn_air_quality,
author = {Chen, Ling and Xu, Jiahui and Wu, Binqing and Huang, Jianlong},
title = {Group-Aware Graph Neural Network for Nationwide City Air Quality Forecasting},
year = {2023},
issue_date = {April 2024},
publisher = {Association for Computing Machinery},
address = {New York, NY, USA},
volume = {18},
number = {3},
issn = {1556-4681},
url = {https://doi.org/10.1145/3631713},
doi = {10.1145/3631713},
journal = {ACM Trans. Knowl. Discov. Data},
month = dec,
articleno = {55},
numpages = {20}
}

@INPROCEEDINGS{arya_2023_hierarchical_stock,
  author={Arya, Arie N. and Lei Xu, Yao and Stankovic, L. and Mandic, Danilo P.},
  booktitle={ICASSP 2023 - 2023 IEEE International Conference on Acoustics, Speech and Signal Processing (ICASSP)}, 
  title={Hierarchical Graph Learning for Stock Market Prediction Via a Domain-Aware Graph Pooling Operator}, 
  year={2023},
  volume={},
  number={},
  pages={1-5},
  doi={10.1109/ICASSP49357.2023.10095381}}

@misc{kipf_2017_gconv,
      title={Semi-Supervised Classification with Graph Convolutional Networks}, 
      author={Thomas N. Kipf and Max Welling},
      year={2017},
      eprint={1609.02907},
      archivePrefix={arXiv},
      primaryClass={cs.LG},
      url={https://arxiv.org/abs/1609.02907}, 
}

@InProceedings{gao_2019_graph_unet,
  title = 	 {Graph U-Nets},
  author =       {Gao, Hongyang and Ji, Shuiwang},
  booktitle = 	 {Proceedings of the 36th International Conference on Machine Learning},
  pages = 	 {2083--2092},
  year = 	 {2019},
  editor = 	 {Chaudhuri, Kamalika and Salakhutdinov, Ruslan},
  volume = 	 {97},
  series = 	 {Proceedings of Machine Learning Research},
  month = 	 {09--15 Jun},
  publisher =    {PMLR},
  url = 	 {https://proceedings.mlr.press/v97/gao19a.html}
}

@ARTICLE{bianchi_2022_hierarchical_gnn,
  author={Bianchi, Filippo Maria and Grattarola, Daniele and Livi, Lorenzo and Alippi, Cesare},
  journal={IEEE Transactions on Neural Networks and Learning Systems}, 
  title={Hierarchical Representation Learning in Graph Neural Networks With Node Decimation Pooling}, 
  year={2022},
  volume={33},
  number={5},
  pages={2195-2207},
  doi={10.1109/TNNLS.2020.3044146}}

@article{bianchi_2022_reservoir,
title = {Pyramidal Reservoir Graph Neural Network},
journal = {Neurocomputing},
volume = {470},
pages = {389-404},
year = {2022},
issn = {0925-2312},
doi = {https://doi.org/10.1016/j.neucom.2021.04.131},
url = {https://www.sciencedirect.com/science/article/pii/S0925231221011152},
author = {F.M. Bianchi and Claudio Gallicchio and Alessio Micheli}
}

@inproceedings{ying_2018_hierarchical_graph,
 author = {Ying, Zhitao and You, Jiaxuan and Morris, Christopher and Ren, Xiang and Hamilton, Will and Leskovec, Jure},
 booktitle = {Advances in Neural Information Processing Systems},
 editor = {S. Bengio and H. Wallach and H. Larochelle and K. Grauman and N. Cesa-Bianchi and R. Garnett},
 pages = {},
 publisher = {Curran Associates, Inc.},
 title = {Hierarchical Graph Representation Learning with Differentiable Pooling},
 url = {https://proceedings.neurips.cc/paper_files/paper/2018/file/e77dbaf6759253c7c6d0efc5690369c7-Paper.pdf},
 volume = {31},
 year = {2018}
}

@inproceedings{
li_2018_diffusion,
title={Diffusion Convolutional Recurrent Neural Network: Data-Driven Traffic Forecasting},
author={Yaguang Li and Rose Yu and Cyrus Shahabi and Yan Liu},
booktitle={International Conference on Learning Representations},
year={2018},
url={https://openreview.net/forum?id=SJiHXGWAZ},
}

@inproceedings{cini_2023_scalable,
author = {Cini, Andrea and Marisca, Ivan and Bianchi, Filippo Maria and Alippi, Cesare},
title = {Scalable spatiotemporal graph neural networks},
year = {2023},
isbn = {978-1-57735-880-0},
publisher = {AAAI Press},
url = {https://doi.org/10.1609/aaai.v37i6.25880},
doi = {10.1609/aaai.v37i6.25880},
booktitle = {Proceedings of the Thirty-Seventh AAAI Conference on Artificial Intelligence and Thirty-Fifth Conference on Innovative Applications of Artificial Intelligence and Thirteenth Symposium on Educational Advances in Artificial Intelligence},
articleno = {811},
numpages = {9},
series = {AAAI'23/IAAI'23/EAAI'23}
}

@inproceedings{cao_2025_stdn,
author = {Cao, Lingxiao and Wang, Bin and Jiang, Guiyuan and Yu, Yanwei and Dong, Junyu},
title = {Spatiotemporal-aware trend-seasonality decomposition network for traffic flow forecasting},
year = {2025},
isbn = {978-1-57735-897-8},
publisher = {AAAI Press},
url = {https://doi.org/10.1609/aaai.v39i11.33247},
doi = {10.1609/aaai.v39i11.33247},
booktitle = {Proceedings of the Thirty-Ninth AAAI Conference on Artificial Intelligence and Thirty-Seventh Conference on Innovative Applications of Artificial Intelligence and Fifteenth Symposium on Educational Advances in Artificial Intelligence},
articleno = {1274},
numpages = {9},
series = {AAAI'25/IAAI'25/EAAI'25}
}

@inproceedings{liu_2023_staeformer,
author = {Liu, Hangchen and Dong, Zheng and Jiang, Renhe and Deng, Jiewen and Deng, Jinliang and Chen, Quanjun and Song, Xuan},
title = {Spatio-Temporal Adaptive Embedding Makes Vanilla Transformer SOTA for Traffic Forecasting},
year = {2023},
isbn = {9798400701245},
publisher = {Association for Computing Machinery},
address = {New York, NY, USA},
url = {https://doi.org/10.1145/3583780.3615160},
doi = {10.1145/3583780.3615160},
booktitle = {Proceedings of the 32nd ACM International Conference on Information and Knowledge Management},
pages = {4125–4129},
numpages = {5},
location = {Birmingham, United Kingdom},
series = {CIKM '23}
}

@inproceedings{
castelli_2026_adaptive,
title={Adaptive Memory Retention in Dynamic Graphs},
author={Fabrizio De Castelli and Alessio Gravina and Moshe Eliasof and Carola-Bibiane Sch{\"o}nlieb and Davide Bacciu},
booktitle={Forty-third International Conference on Machine Learning},
year={2026},
url={https://openreview.net/forum?id=x7NGYPNlgs}
}

@inproceedings{zheng_2015_forecasting,
author = {Zheng, Yu and Yi, Xiuwen and Li, Ming and Li, Ruiyuan and Shan, Zhangqing and Chang, Eric and Li, Tianrui},
title = {Forecasting Fine-Grained Air Quality Based on Big Data},
year = {2015},
isbn = {9781450336642},
publisher = {Association for Computing Machinery},
address = {New York, NY, USA},
url = {https://doi.org/10.1145/2783258.2788573},
doi = {10.1145/2783258.2788573},
booktitle = {Proceedings of the 21th ACM SIGKDD International Conference on Knowledge Discovery and Data Mining},
pages = {2267–2276},
numpages = {10},
location = {Sydney, NSW, Australia},
series = {KDD '15}
}

@misc{Zippenfenig_Open-Meteo,
  author    = {Zippenfenig, Patrick},
  doi       = {10.5281/zenodo.7970649},
  license   = {CC-BY-4.0},
  title     = {Open-Meteo.com Weather API},
  year      = {2023},
  copyright = {Creative Commons Attribution 4.0 International},
  url       = {https://open-meteo.com/}
}

@inproceedings{cini_2022_filling,
  author={Andrea Cini and Ivan Marisca and Cesare Alippi},
  title={Filling the {G\_ap\_s}: Multivariate Time Series Imputation by Graph Neural Networks},
  year={2022},
  cdate={1640995200000},
  url={https://openreview.net/forum?id=kOu3-S3wJ7},
  booktitle={International Conference on Learning Representations}
}

@inproceedings{liu_2023_largest,
author = {Liu, Xu and Xia, Yutong and Liang, Yuxuan and Hu, Junfeng and Wang, Yiwei and Bai, Lei and Huang, Chao and Liu, Zhenguang and Hooi, Bryan and Zimmermann, Roger},
title = {LargeST: a benchmark dataset for large-scale traffic forecasting},
year = {2023},
publisher = {Curran Associates Inc.},
address = {Red Hook, NY, USA},
booktitle = {Proceedings of the 37th International Conference on Neural Information Processing Systems},
articleno = {3293},
numpages = {18},
location = {New Orleans, LA, USA},
series = {NIPS '23}
}

@inproceedings{yi_2016_st,
author = {Yi, Xiuwen and Zheng, Yu and Zhang, Junbo and Li, Tianrui},
title = {ST-MVL: filling missing values in geo-sensory time series data},
year = {2016},
isbn = {9781577357704},
publisher = {AAAI Press},
booktitle = {Proceedings of the Twenty-Fifth International Joint Conference on Artificial Intelligence},
pages = {2704–2710},
numpages = {7},
location = {New York, New York, USA},
series = {IJCAI'16}
}

@incollection{paske2019pytorch,
  author    = {Paszke, Adam and Gross, Sam and Massa, Francisco and
               Lerer, Adam and Bradbury, James and Chanan, Gregory and
               Gimelshein, Natalia and Antiga, Luca and
               Desmaison, Alban and Kopf, Andreas and Yang, Edward and
               DeVito, Zachary and Raison, Martin and
               Tejani, Alykhan and Chilamkurthy, Sasank and
               Steiner, Benoit and Fang, Lu and Bai, Junjie and
               Chintala, Soumith},
  booktitle = {Advances in Neural Information Processing Systems 32},
  editor    = {H. Wallach and H. Larochelle and A. Beygelzimer and
               F. d\textquotesingle Alch\'{e}-Buc and E. Fox and
               R. Garnett},
  pages     = {8024--8035},
  publisher = {Curran Associates, Inc.},
  title     = {PyTorch: An Imperative Style, High-Performance Deep
               Learning Library},
  year      = {2019}
}

@inproceedings{fey2019fast,
  title={Fast Graph Representation Learning with {PyTorch Geometric}},
  author={Fey, Matthias and Lenssen, Jan E.},
  booktitle={ICLR Workshop on Representation Learning on Graphs and Manifolds},
  year={2019},
}

@article{harris2020array,
  author    = {Harris, Charles R and Millman, K Jarrod and
               Van Der Walt, St{\'e}fan J and Gommers, Ralf and
               Virtanen, Pauli and Cournapeau, David and
               Wieser, Eric and Taylor, Julian and Berg, Sebastian and
               Smith, Nathaniel J and others},
  journal   = {Nature},
  number    = {7825},
  pages     = {357--362},
  publisher = {Nature Publishing Group},
  title     = {Array programming with NumPy},
  volume    = {585},
  year      = {2020}
}

@misc{Cini_Torch_Spatiotemporal_2022,
    author = {Cini, Andrea and Marisca, Ivan},
    license = {MIT},
    month = {3},
    title = {{Torch Spatiotemporal}},
    url = {https://github.com/TorchSpatiotemporal/tsl},
    year = {2022}
}

@misc{Falcon_PyTorch_Lightning_2019,
  author  = {Falcon, William and {The PyTorch Lightning team}},
  doi     = {10.5281/zenodo.3828935},
  license = {Apache-2.0},
  month   = {3},
  title   = {{PyTorch Lightning}},
  url     = {https://github.com/PyTorchLightning/pytorch-lightning},
  version = {1.4},
  year    = {2019}
}

@Misc{Yadan2019Hydra,
  author =       {Omry Yadan},
  title =        {Hydra - A framework for elegantly configuring complex applications},
  howpublished = {Github},
  year =         {2019},
  url =          {https://github.com/facebookresearch/hydra}
}

@book{rossum2009python,
 author = {Van Rossum, Guido and Drake, Fred L.},
 title = {Python 3 Reference Manual},
 year = {2009},
 isbn = {1441412697},
 publisher = {CreateSpace},
 address = {Scotts Valley, CA}
}

@InProceedings{rangapuram_2023_coherent,
  title = 	 {Coherent Probabilistic Forecasting of Temporal Hierarchies},
  author =       {Rangapuram, Syama Sundar and Kapoor, Shubham and Nirwan, Rajbir Singh and Mercado, Pedro and Januschowski, Tim and Wang, Yuyang and Bohlke-Schneider, Michael},
  booktitle = 	 {Proceedings of The 26th International Conference on Artificial Intelligence and Statistics},
  pages = 	 {9362--9376},
  year = 	 {2023},
  editor = 	 {Ruiz, Francisco and Dy, Jennifer and van de Meent, Jan-Willem},
  volume = 	 {206},
  series = 	 {Proceedings of Machine Learning Research},
  month = 	 {25--27 Apr},
  publisher =    {PMLR},
  url = 	 {https://proceedings.mlr.press/v206/rangapuram23a.html}
}

@article{athanasopoulos_2017_forecasting,
title = {Forecasting with temporal hierarchies},
journal = {European Journal of Operational Research},
volume = {262},
number = {1},
pages = {60-74},
year = {2017},
issn = {0377-2217},
doi = {https://doi.org/10.1016/j.ejor.2017.02.046},
url = {https://www.sciencedirect.com/science/article/pii/S0377221717301911},
author = {George Athanasopoulos and Rob J. Hyndman and Nikolaos Kourentzes and Fotios Petropoulos}
}
\bibliographystyle{main}

\appendix
\section{Data}
\label{a:data}
\begin{table*}[ht]

\caption{Overview of the datasets.}
\label{t:datasets}

\vspace{0.1in}
\setlength{\tabcolsep}{4pt}
\renewcommand{\arraystretch}{1.2}
\centering
\begin{small}
\begin{tabular}{l|c c c c c c || c c}
\toprule
 \sc Datasets & Time steps & Nodes & Edges & Type & Rate & Channels & Window & Horizon\\
\toprule
\gls{la} & 34,272 & 207 & 1,515 & Directed  & 5 minutes & 1 & 12 & 12\\
\gls{bay} & 52,128 & 325 & 2,369 & Directed & 5 minutes & 1 & 12 & 12\\
\gls{air} & 8,760 & 437 & 2,699 & Undirected & 1 hour & 1 & 24 & 3\\
\gls{engrad} & 26,304 & 487 & 2,297 & Undirected & 1 hour & 5 & 24 & 6\\
\gls{gla} & 35,040 & 3,834 & 98,703 & Directed & 15 minutes & 1 & 96 & 12 \\
\bottomrule
\end{tabular}
\end{small}
\end{table*} 
We use the following datasets:
\begin{itemize}[leftmargin=1.5em]
    \item \textbf{\gls{la}}~\citep{li_2018_diffusion} contains traffic measurements recorded by loop detectors on highways in Los Angeles County. 
    \item \textbf{\gls{bay}}~\citep{li_2018_diffusion} reports traffic speeds observed by road sensors located throughout the San Francisco Bay Area.
    \item \textbf{\gls{air}}~\citep{zheng_2015_forecasting} records PM2.5 concentrations at an hourly resolution through $437$ stations distributed across $43$ cities in China. 
    \item \textbf{\gls{engrad}} contains $3$ years of $5$ historical weather variables sampled hourly at $487$ grid points in England. The measurements are provided by \href{https://open-meteo.com/}{open-meteo.com}~\citep{Zippenfenig_Open-Meteo,marisca_2024_graph_missing_downsampling}.
\end{itemize}

We provide details about the datasets in~\autoref{t:datasets}. Following previous work, we use the standard adjacency matrix and exogenous variables associated with each dataset~\citep{cini_2022_filling, li_2018_diffusion, cini_2023_taming,marisca_2024_graph_missing_downsampling}. 
We partition each time series into training, validation, and test sets, containing $70\%/10\%/20\%$ of the observations, respectively. The only exception is \gls{air}, for which we adopt the predefined splits established in the literature~\citep{yi_2016_st}.
\begin{figure}[t]
    \centering
    \includegraphics[width=\linewidth]{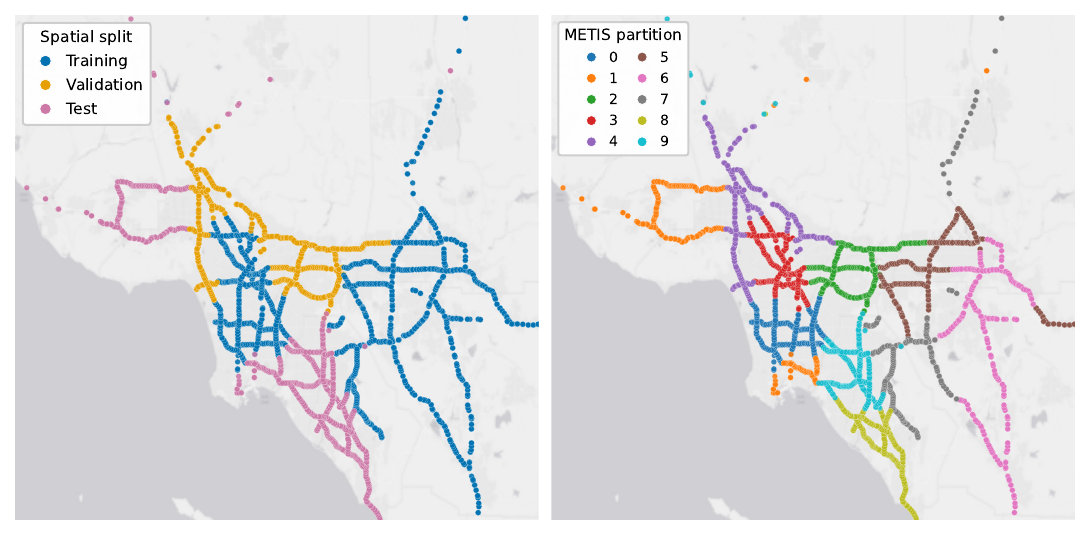}
    \caption{Spatial partitioning of the \gls{gla} sensor graph. The left panel shows the nodes used for training, validation, and test, whereas the right panel shows the ten partitions. A small number of distant sensors fall outside the displayed map and are omitted for visualization only.}
    \label{fig:largest-inductive-split}
\end{figure}
Additionally, for the inductive setting, we use \textbf{\gls{largest}}~\citep{liu_2023_largest}, a large-scale traffic forecasting benchmark based on speed measurements from PeMS sensors across California. We consider the \gls{gla} subset, which contains $3{,}834$ sensors connected by $98{,}703$ directed and weighted road-network edges. We use the 2019 observations, aggregated at a 15-minute frequency. We partition the \gls{gla} road graph into ten balanced and disjoint subgraphs.
Edges between different partitions are removed, and each partition is subsequently treated as an independent spatiotemporal graph. The resulting partitions contain between 372 and 394 nodes. We assign five subgraphs to training, two subgraphs to validation, and the remaining three subgraphs to test (see \autoref{tab:largest-inductive-partitions} and \autoref{fig:largest-inductive-split}). Within each subgraph, we apply the same $70\%/10\%/20\%$ chronological split. At test time, we evaluate the model’s ability to transfer inductively across graph instances with disjoint node sets and different connectivity patterns. A separate standard scaler is associated with each subgraph and is estimated from its training data.

\begin{table}[t]
\centering
\caption{\gls{gla} spatial split. The left table shows the assignment of the ten partitions to the training, validation, and test sets, while the right table reports detailed statistics for each partition. \\
}

\label{tab:largest-inductive-partitions}
\begin{minipage}[t]{0.48\textwidth}
\centering
\vspace{0pt}
\begin{tabular}{lccr}
\toprule
Split & Partitions & No. & Sensors \\
\midrule
Training
    & $\{5,6,0,7,3\}$ & 5 & 1,932 \\
Validation
    & $\{2,4\}$       & 2 &   769 \\
Test
    & $\{1,8,9\}$     & 3 & 1,133 \\
\midrule
Total
    & $\{0,\ldots,9\}$ & 10 & 3,834 \\
\bottomrule
\end{tabular}
\end{minipage}
\hfill
\begin{minipage}[t]{0.48\textwidth}
\centering
\vspace{0pt}
\begin{tabular}{clrr}
\toprule
Partition & Split & Nodes & Edges \\
\midrule
0 & Training   & 382 & 12,522  \\
1 & Test       & 372 &  5,714  \\
2 & Validation & 394 &  9,786  \\
3 & Training   & 391 & 15,451  \\
4 & Validation & 375 &  8,086  \\
5 & Training   & 380 &  6,393 \\
6 & Training   & 393 &  5,800  \\
7 & Training   & 386 &  4,728  \\
8 & Test       & 372 & 11,642 \\
9 & Test       & 389 & 14,506 \\
\midrule
\multicolumn{2}{l}{Total}
    & 3,834 & 94,628 \\
\bottomrule
\end{tabular}
\end{minipage}
\end{table}

\section{Baselines}
\label{a:baselines}
In the experiments, we use the following baselines:
\begin{itemize}
    \item \textbf{\Gls{rnn}}, a global recurrent model that processes each node independently with shared parameters, without using the graph structure;
    \item \textbf{\Gls{dcrnn}}~\citep{li_2018_diffusion} models temporal dynamics with a \gls{rnn} and uses diffusion graph convolutions to exchange information among nodes;
    \item \textbf{GConv-TTS}~\citep{kipf_2017_gconv} first encodes each series with a \gls{rnn} and then applies graph convolutions to exchange information among neighboring nodes;
    \item \textbf{GUNet-TTS}~\citep{gao_2019_graph_unet} first encodes each series with a \gls{rnn} and then uses graph pooling and unpooling to process the resulting representations at multiple spatial resolutions;
    \item \textbf{Graph WaveNet} (\textbf{GWNet})~\citep{wu_2019_graph_wavenet} combines dilated temporal convolutions with graph convolutions;
    \item \textbf{SGP}~\citep{cini_2023_scalable} first encodes temporal and spatial information using a randomized spatiotemporal graph encoder and then maps the resulting representations to forecasts with a trainable model;
    \item \textbf{STAEformer}~\citep{liu_2023_staeformer} uses a Transformer with learned spatial and temporal embeddings;
    \item \textbf{STDN}~\citep{cao_2025_stdn} learns dynamic graph representations and decomposes the temporal signals into trend and seasonal components;
    \item \textbf{LAMP}~\citep{castelli_2026_adaptive} models the evolution of node representations through stable graph dynamics. Its adaptive dissipation mechanism controls how much past information is retained or discarded, supporting the propagation of information over long spatial and temporal ranges;
    \item \textbf{\gls{higp}}~\citep{cini_2024_higp} learns a hierarchy from the input graph and produces coherent forecasts at each hierarchical level. Its \gls{mp} scheme combines graph convolutions at the bottom level with bidirectional propagation between adjacent hierarchy levels through the learned aggregation matrices.
\end{itemize}
\gls{higp}, GConv-TTS, and GUNet-TTS share the same \gls{tts} pipeline \citep{gao_2021_equivalence}. They encode the temporal history using an \gls{rnn}, optionally use learnable node embeddings \citep{cini_2023_taming}, apply model-specific \gls{mp}, and produce forecasts with an \gls{mlp} readout. They differ in how spatial information is propagated, with \gls{higp} additionally processing representations across multiple hierarchical levels. We adopt the same pipeline in \gls{shr}. In the inductive setting, we disable node embeddings for all models, as embeddings learned on the training graph cannot be transferred to unseen series in the test graph.

\section{Experimental Details and Additional Results}
\label{a:additional_results}
This section provides additional details on the experimental setup and supplementary results. Throughout the paper, all models are trained using the \gls{mae} loss. Results reported in tables and figures are averaged over at least three independent runs with different random seeds. 

\begin{table*}[htbp]

\caption{Performance across datasets at selected forecasting horizons. \textbf{Bold} and \underline{underlined} values denote the best and second-best \gls{mae}, respectively.}
\label{tab:baselines_steps}

\centering
\resizebox{\textwidth}{!}{
\begin{tabular}{l|rrr|rrr|rrr|rrr}
\toprule
Model & \multicolumn{3}{c}{\gls{la}} & \multicolumn{3}{c}{\gls{bay}} & \multicolumn{3}{c}{\gls{air}} & \multicolumn{3}{c}{\gls{engrad}} \\
\cmidrule(lr){2-4} \cmidrule(lr){5-7} \cmidrule(lr){8-10} \cmidrule(lr){11-13}
 & @3 & @6 & @12 & @3 & @6 & @12 & @1 & @2 & @3 & @2 & @3 & @6 \\
\midrule
RNN & 2.977 {\tiny $\pm\,.001$} & 3.556 {\tiny $\pm\,.001$} & 4.356 {\tiny $\pm\,.004$} & 1.389 {\tiny $\pm\,.000$} & 1.815 {\tiny $\pm\,.001$} & 2.307 {\tiny $\pm\,.001$} & 9.783 {\tiny $\pm\,.053$} & 14.310 {\tiny $\pm\,.051$} & 17.665 {\tiny $\pm\,.052$} & 37.707 {\tiny $\pm\,.272$} & 46.570 {\tiny $\pm\,.480$} & 60.928 {\tiny $\pm\,.819$} \\
DCRNN & 2.865 {\tiny $\pm\,.001$} & 3.319 {\tiny $\pm\,.003$} & 3.871 {\tiny $\pm\,.003$} & 1.372 {\tiny $\pm\,.002$} & 1.759 {\tiny $\pm\,.003$} & 2.184 {\tiny $\pm\,.006$} & 9.058 {\tiny $\pm\,.011$} & 12.958 {\tiny $\pm\,.019$} & 16.039 {\tiny $\pm\,.030$} & 34.025 {\tiny $\pm\,.057$} & 42.827 {\tiny $\pm\,.068$} & 57.952 {\tiny $\pm\,.156$} \\
GConv-TTS & 2.764 {\tiny $\pm\,.008$} & 3.122 {\tiny $\pm\,.004$} & 3.500 {\tiny $\pm\,.014$} & 1.321 {\tiny $\pm\,.002$} & 1.633 {\tiny $\pm\,.008$} & 1.907 {\tiny $\pm\,.012$} & 9.055 {\tiny $\pm\,.050$} & 12.603 {\tiny $\pm\,.017$} & 15.522 {\tiny $\pm\,.028$} & 34.402 {\tiny $\pm\,.211$} & 42.929 {\tiny $\pm\,.401$} & 57.735 {\tiny $\pm\,.173$} \\
GUNet-TTS & 2.748 {\tiny $\pm\,.003$} & 3.101 {\tiny $\pm\,.003$} & 3.489 {\tiny $\pm\,.002$} & 1.322 {\tiny $\pm\,.006$} & 1.633 {\tiny $\pm\,.006$} & 1.905 {\tiny $\pm\,.002$} & 9.029 {\tiny $\pm\,.012$} & 12.547 {\tiny $\pm\,.019$} & 15.401 {\tiny $\pm\,.043$} & 35.099 {\tiny $\pm\,.583$} & 43.689 {\tiny $\pm\,.609$} & 58.640 {\tiny $\pm\,1.359$} \\
GWNet & 2.745 {\tiny $\pm\,.013$} & 3.120 {\tiny $\pm\,.002$} & 3.514 {\tiny $\pm\,.009$} & 1.317 {\tiny $\pm\,.003$} & 1.644 {\tiny $\pm\,.009$} & 1.914 {\tiny $\pm\,.013$} & \textbf{8.719 {\tiny $\pm\,.041$}} & 12.277 {\tiny $\pm\,.036$} & \underline{14.891 {\tiny $\pm\,.026$}} & 32.760 {\tiny $\pm\,.477$} & \underline{39.598 {\tiny $\pm\,.289$}} & 53.373 {\tiny $\pm\,.677$} \\
SGP & 2.750 {\tiny $\pm\,.002$} & 3.104 {\tiny $\pm\,.003$} & 3.460 {\tiny $\pm\,.002$} & 1.346 {\tiny $\pm\,.004$} & 1.665 {\tiny $\pm\,.004$} & 1.932 {\tiny $\pm\,.005$} & 9.120 {\tiny $\pm\,.011$} & 13.023 {\tiny $\pm\,.011$} & 16.041 {\tiny $\pm\,.009$} & 34.227 {\tiny $\pm\,.163$} & 43.046 {\tiny $\pm\,.216$} & 59.052 {\tiny $\pm\,.069$} \\
STAEformer & \underline{2.722 {\tiny $\pm\,.028$}} & \underline{3.056 {\tiny $\pm\,.032$}} & \underline{3.451 {\tiny $\pm\,.003$}} & \underline{1.310 {\tiny $\pm\,.007$}} & \textbf{1.615 {\tiny $\pm\,.013$}} & \textbf{1.881 {\tiny $\pm\,.005$}} & 9.525 {\tiny $\pm\,.094$} & 13.121 {\tiny $\pm\,.061$} & 15.787 {\tiny $\pm\,.062$} & 47.982 {\tiny $\pm\,.811$} & 57.277 {\tiny $\pm\,1.064$} & 71.587 {\tiny $\pm\,1.009$} \\
LAMP & 2.853 {\tiny $\pm\,.001$} & 3.174 {\tiny $\pm\,.002$} & 3.515 {\tiny $\pm\,.004$} & 1.477 {\tiny $\pm\,.002$} & 1.734 {\tiny $\pm\,.002$} & 1.968 {\tiny $\pm\,.003$} & 12.772 {\tiny $\pm\,.040$} & 16.043 {\tiny $\pm\,.027$} & 18.668 {\tiny $\pm\,.018$} & 43.072 {\tiny $\pm\,.153$} & 49.473 {\tiny $\pm\,.142$} & 61.071 {\tiny $\pm\,.143$} \\
STDN & 2.767 {\tiny $\pm\,.012$} & 3.095 {\tiny $\pm\,.006$} & 3.457 {\tiny $\pm\,.019$} & 1.335 {\tiny $\pm\,.002$} & 1.649 {\tiny $\pm\,.006$} & 1.914 {\tiny $\pm\,.010$} & 10.212 {\tiny $\pm\,.248$} & 13.399 {\tiny $\pm\,.165$} & 16.056 {\tiny $\pm\,.185$} & 43.284 {\tiny $\pm\,.786$} & 51.793 {\tiny $\pm\,.551$} & 67.643 {\tiny $\pm\,.262$} \\
HiGP & 2.734 {\tiny $\pm\,.010$} & 3.077 {\tiny $\pm\,.011$} & 3.456 {\tiny $\pm\,.006$} & 1.333 {\tiny $\pm\,.001$} & 1.644 {\tiny $\pm\,.002$} & 1.917 {\tiny $\pm\,.008$} & 8.800 {\tiny $\pm\,.010$} & \underline{12.242 {\tiny $\pm\,.028$}} & 14.911 {\tiny $\pm\,.035$} & \underline{32.350 {\tiny $\pm\,.405$}} & 40.117 {\tiny $\pm\,.518$} & \underline{53.055 {\tiny $\pm\,.435$}} \\
\midrule
SHR & \textbf{2.694 {\tiny $\pm\,.015$}} & \textbf{3.046 {\tiny $\pm\,.016$}} & \textbf{3.424 {\tiny $\pm\,.022$}} & \textbf{1.308 {\tiny $\pm\,.007$}} & \underline{1.620 {\tiny $\pm\,.010$}} & \underline{1.888 {\tiny $\pm\,.008$}} & \underline{8.786 {\tiny $\pm\,.013$}} & \textbf{12.215 {\tiny $\pm\,.008$}} & \textbf{14.836 {\tiny $\pm\,.009$}} & \textbf{31.069 {\tiny $\pm\,.200$}} & \textbf{38.259 {\tiny $\pm\,.024$}} & \textbf{51.898 {\tiny $\pm\,.090$}} \\
\bottomrule
\end{tabular}
}
\end{table*}

\subsection{Forecasting performance and hierarchy variants}
For the experiments in \autoref{tab:baselines}, we implemented the baselines in our environment using their original implementations. We generally followed the recommended hyperparameter settings and tuned selected model parameters. 
The batch size and learning rate were set to $32$ and $0.003$, respectively, or to $16$ and $0.001$ in some cases. For \gls{shr}, we tuned the hidden dimension following the same procedure used for the baselines.
Its bottom-level predictions at test time are obtained by averaging forecasts over $20$ independently sampled hierarchies generated using \gls{kmis}.
For \textit{Fixed hier.}, instead, a single hierarchy obtained through spectral clustering is fixed throughout the forward pass, including input aggregation, hierarchical message passing, and forecast reconciliation.
Finally, for \textit{\gls{shr} (rand.)}, we replace the hierarchy construction used by \gls{shr} with random balanced assignments. At each training batch, a new hierarchy is generated by sampling $N_1$ uniformly from $\{10,\ldots,30\}$ and assigning the bottom-level nodes to balanced intermediate groups. At validation, forecasts are averaged over three hierarchies per batch, with $N_1\in\{10,20,30\}$. At test time, they are averaged over $20$ independently generated hierarchies per batch, with $N_1$ sampled uniformly from $\{10,\ldots,30\}$. 
In \autoref{tab:baselines_steps}, we further report forecasting errors at specific forecast horizons. The results are broadly consistent with those in \autoref{tab:baselines}.

\subsection{Sensitivity analysis}
\label{par:ablation_plot}

\begin{figure}[!t]
    \centering
    \captionsetup[subfigure]{font=footnotesize,skip=2pt}

    \begin{subfigure}[t]{0.245\textwidth}
        \centering
        \includegraphics[width=\linewidth]
        {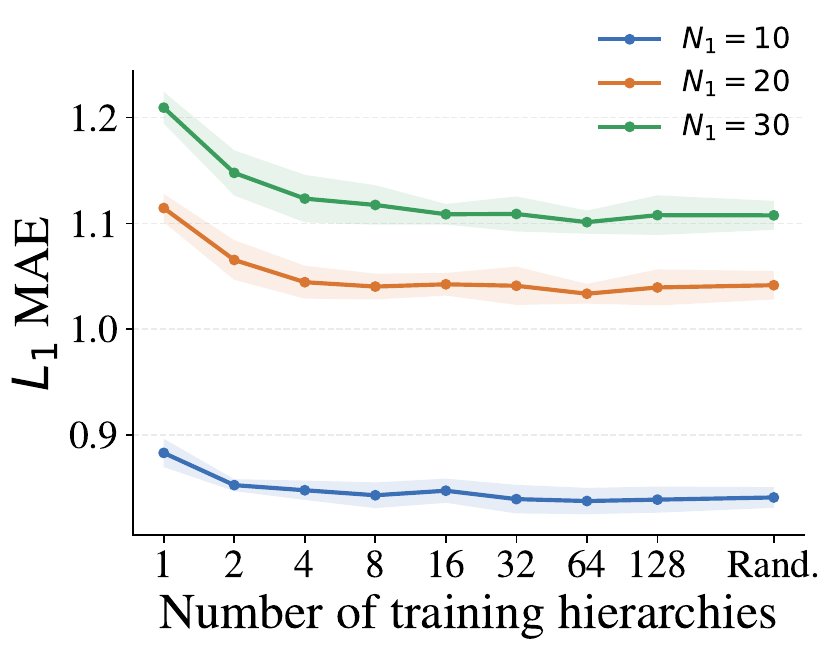}
        \caption{Fixed $N_1$ ($L_1$).}
        \label{fig:hierarchy_count_same_val_test_bay_five_seeds_shadow_L1}
    \end{subfigure}
    \hfill
    \begin{subfigure}[t]{0.245\textwidth}
        \centering
        \includegraphics[width=\linewidth]
        {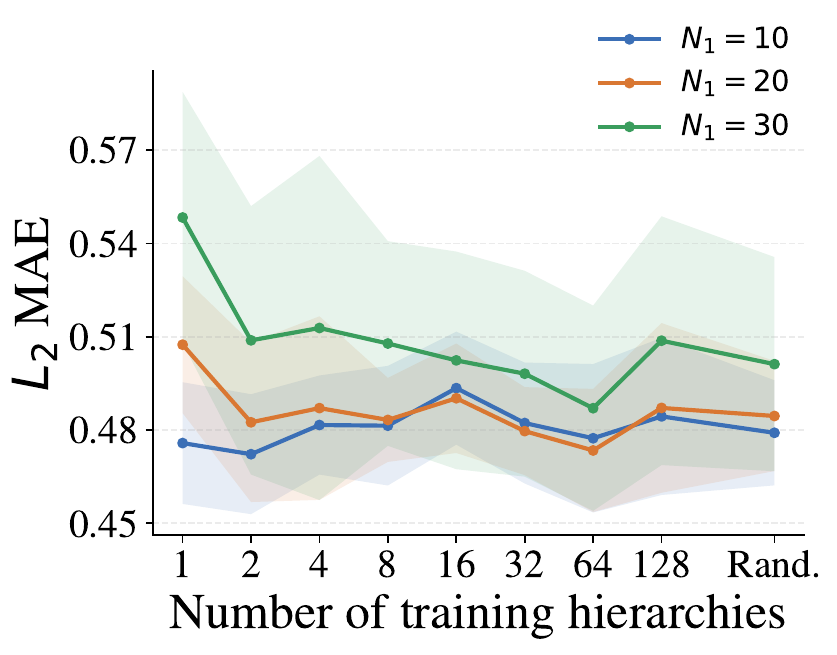}
        \caption{Fixed $N_1$ ($L_2$).}
        \label{fig:hierarchy_count_same_val_test_bay_five_seeds_shadow_L2}
    \end{subfigure}
    \hfill
    \begin{subfigure}[t]{0.245\textwidth}
        \centering
        \includegraphics[width=\linewidth]
        {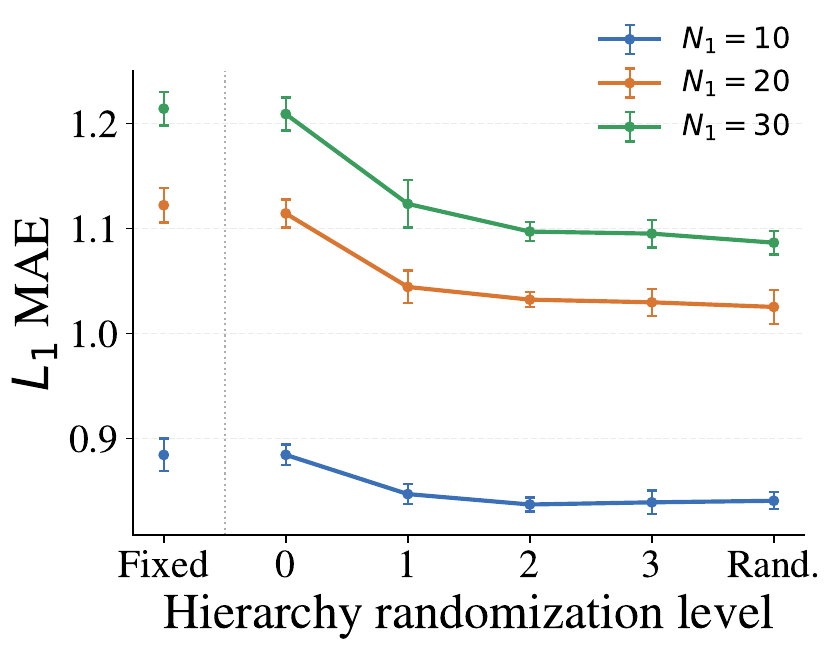}
        \caption{Variable $N_1$ ($L_1$).}
        \label{fig:progressive_same_val_test_bay_five_seeds_L1}
    \end{subfigure}
    \hfill
    \begin{subfigure}[t]{0.245\textwidth}
        \centering
        \includegraphics[width=\linewidth]
        {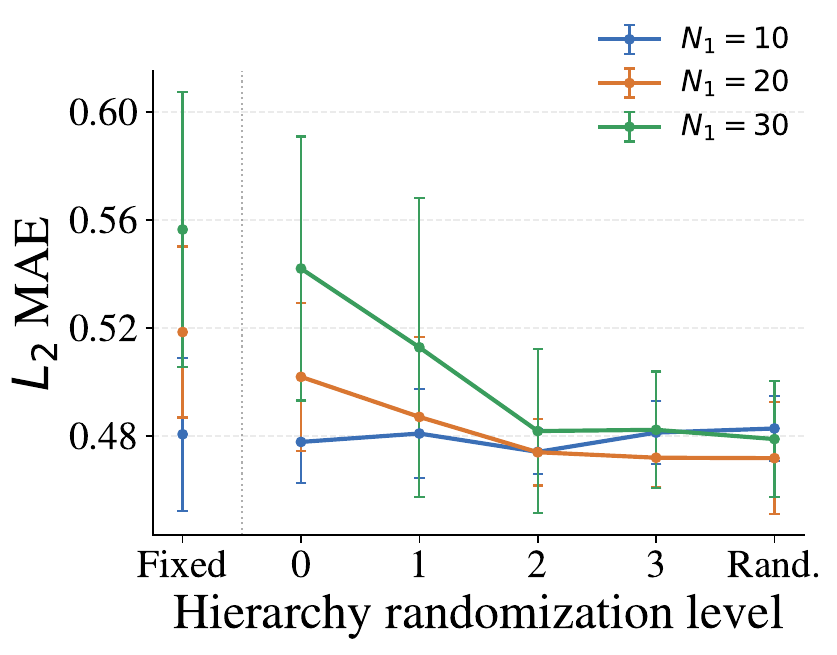}
        \caption{Variable $N_1$ ($L_2$).}
        \label{fig:progressive_same_val_test_bay_five_seeds_L2}
    \end{subfigure}
    \par\vspace{2mm}

    \begin{subfigure}[t]{0.245\textwidth}
        \centering
        \includegraphics[width=\linewidth]
        {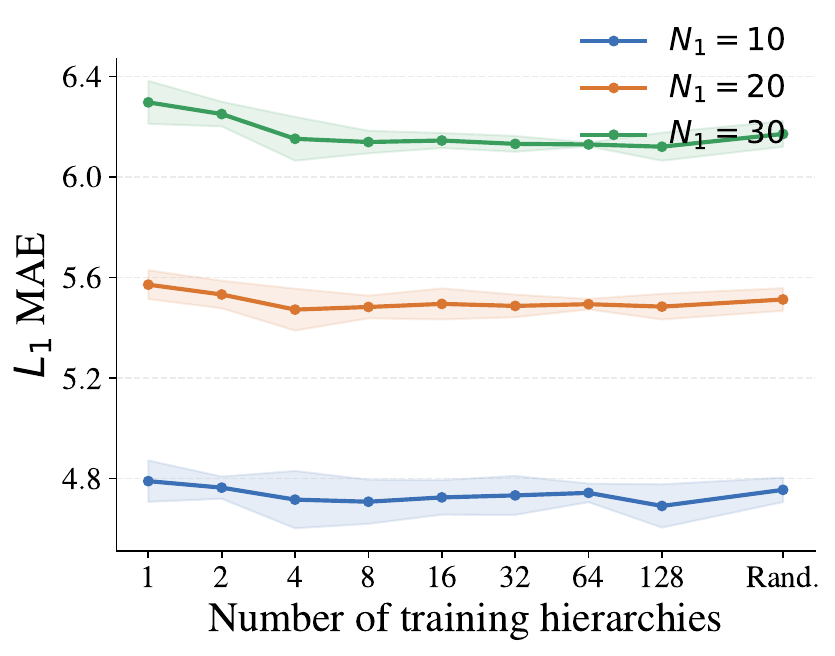}
        \caption{Fixed $N_1$ ($L_1$).}
        \label{fig:hierarchy_count_same_val_test_air_five_seeds_shadow_L1}
    \end{subfigure}
    \hfill
    \begin{subfigure}[t]{0.245\textwidth}
        \centering
        \includegraphics[width=\linewidth]
        {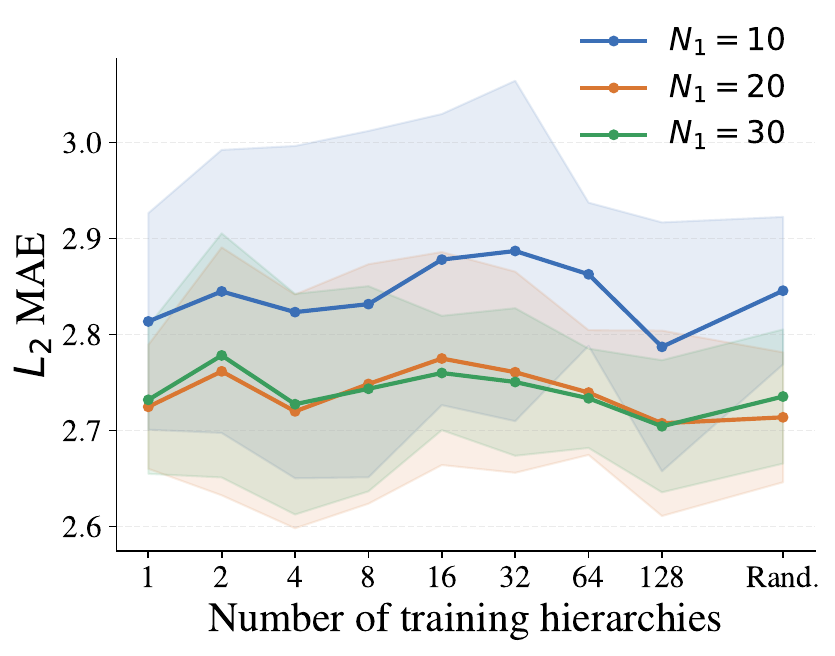}
        \caption{Fixed $N_1$ ($L_2$).}
        \label{fig:hierarchy_count_same_val_test_air_five_seeds_shadow_L2}
    \end{subfigure}
    \hfill
    \begin{subfigure}[t]{0.245\textwidth}
        \centering
        \includegraphics[width=\linewidth]
        {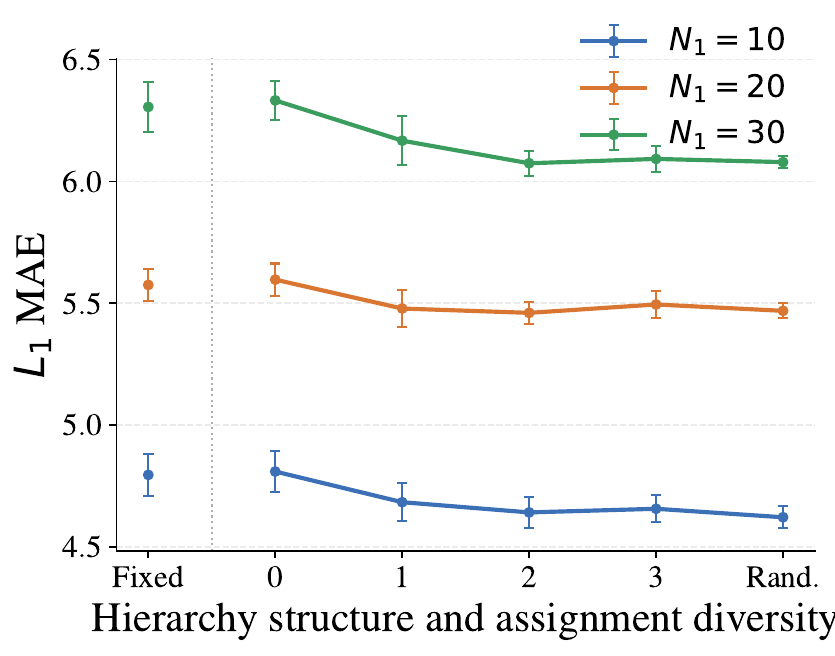}
        \caption{Variable $N_1$ ($L_1$).}
        \label{fig:progressive_same_val_test_air_five_seeds_L1}
    \end{subfigure}
    \hfill
    \begin{subfigure}[t]{0.245\textwidth}
        \centering
        \includegraphics[width=\linewidth]
        {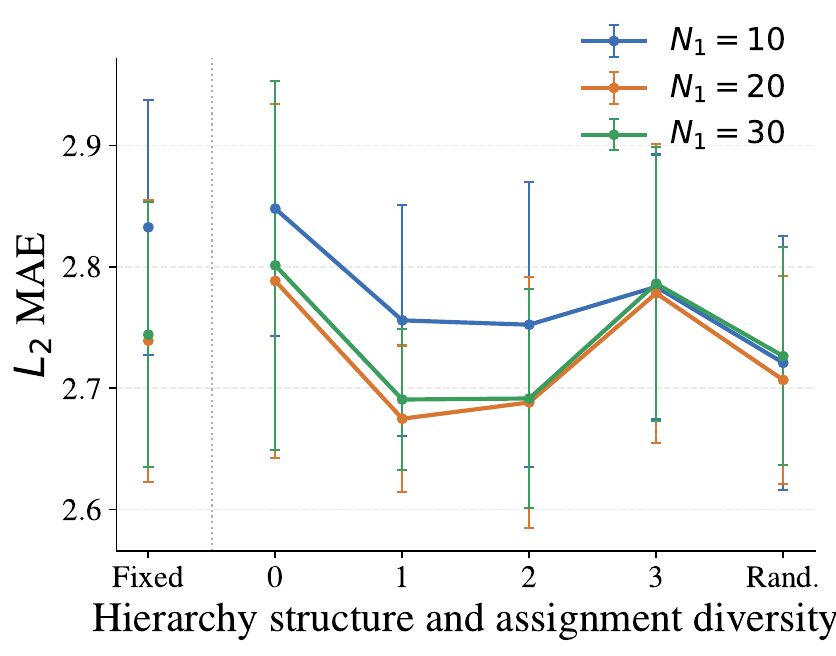}
        \caption{Variable $N_1$ ($L_2$).}
        \label{fig:progressive_same_val_test_air_five_seeds_L2}
    \end{subfigure}

    \par\vspace{2mm}

    \begin{subfigure}[t]{0.245\textwidth}
        \centering
        \includegraphics[width=\linewidth]
        {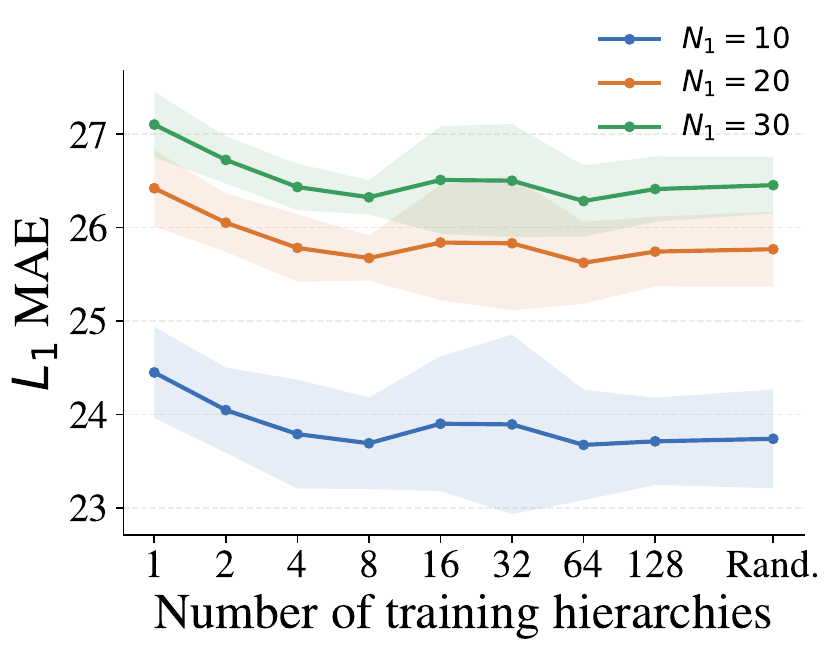}
        \caption{Fixed $N_1$ ($L_1$).}
        \label{fig:hierarchy_count_same_val_test_engrad_five_seeds_shadow_L1}
    \end{subfigure}
    \hfill
    \begin{subfigure}[t]{0.245\textwidth}
        \centering
        \includegraphics[width=\linewidth]
        {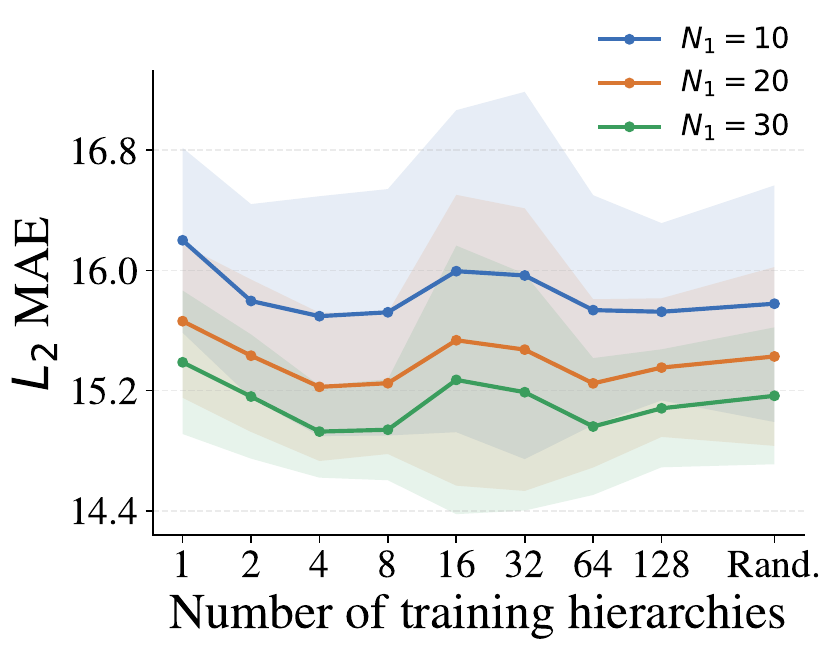}
        \caption{Fixed $N_1$ ($L_2$).}
        \label{fig:hierarchy_count_same_val_test_engrad_five_seeds_shadow_L2}
    \end{subfigure}
    \hfill
    \begin{subfigure}[t]{0.245\textwidth}
        \centering
        \includegraphics[width=\linewidth]
        {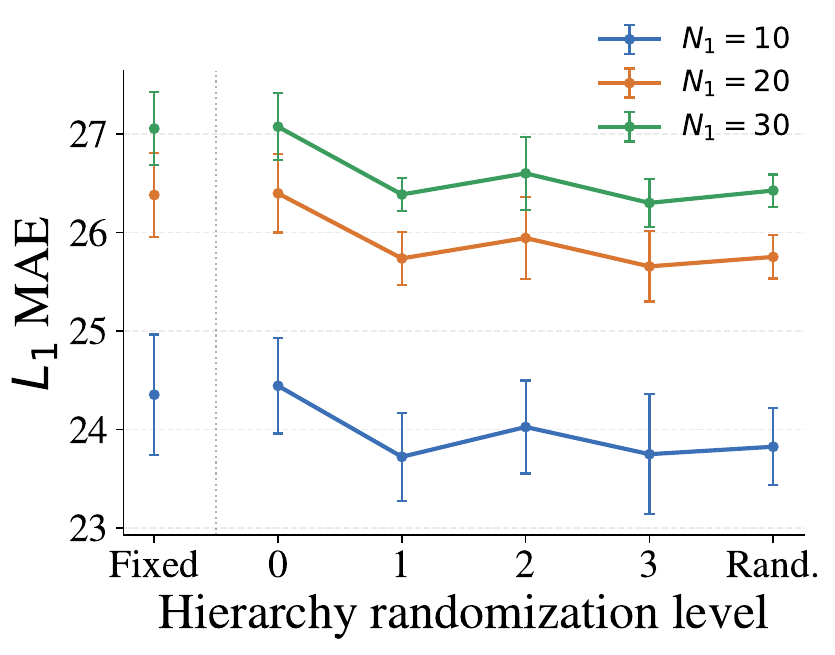}
        \caption{Variable $N_1$ ($L_1$).}
        \label{fig:progressive_same_val_test_engrad_five_seeds_L1}
    \end{subfigure}
    \hfill
    \begin{subfigure}[t]{0.245\textwidth}
        \centering
        \includegraphics[width=\linewidth]
        {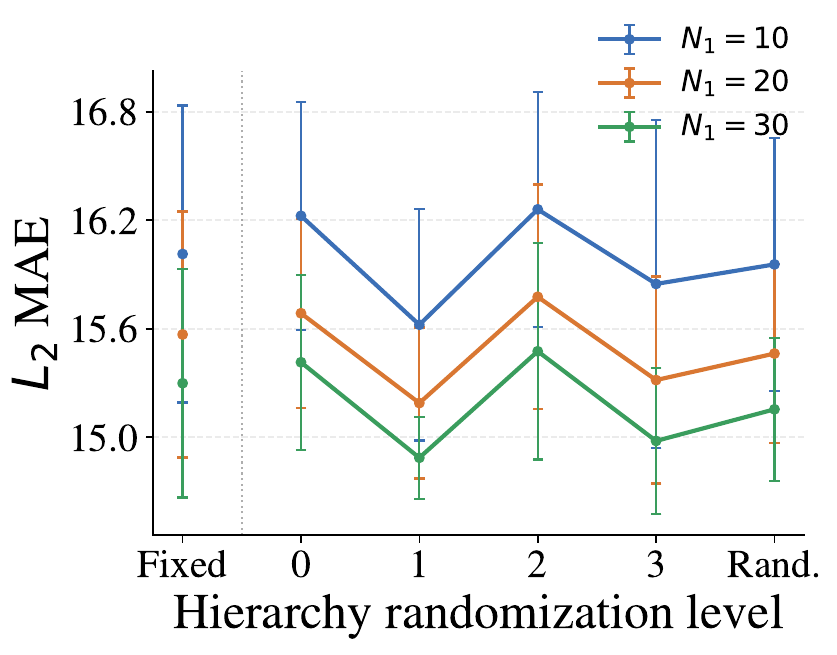}
        \caption{Variable $N_1$ ($L_2$).}
        \label{fig:progressive_same_val_test_engrad_five_seeds_L2}
    \end{subfigure}

    \caption{Performance at higher aggregation levels ($L_1$ and $L_2$) on \gls{bay} (top row), \gls{air} (middle row), and \gls{engrad} (bottom row). The first two columns increase the number of training hierarchies while keeping $N_1=20$ and varying node-to-aggregate assignments. The third and fourth columns instead progressively increase training-hierarchy diversity by varying both the assignments and $N_1$. Each curve evaluates the same set of trained models using common test hierarchies with $N_1\in\{10,20,30\}$.
    }
    \label{fig:hierarchy_variability_higher_levels_datasets}
\end{figure}

\begin{figure}[!t]
    \centering
    \captionsetup[subfigure]{font=footnotesize,skip=2pt}
    \begin{subfigure}[b]{0.325\textwidth}
        \centering
        \includegraphics[width=\linewidth]{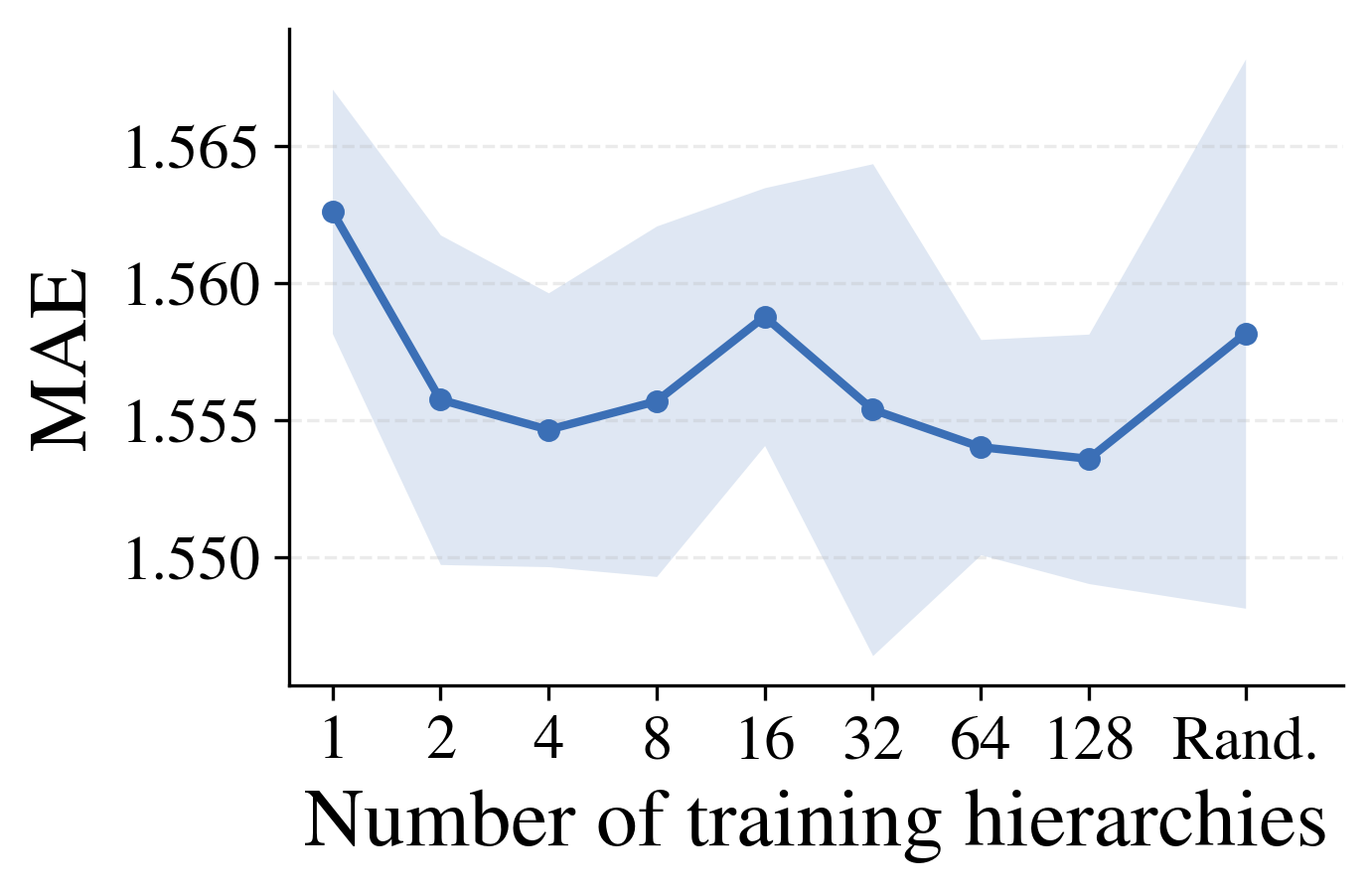}
        \caption{Fixed $N_1$, performance at $L_0$.}
        \label{fig:hierarchy_count_same_val_test_bay_five_seeds}
    \end{subfigure}
    \hfill
    \begin{subfigure}[b]{0.325\textwidth}
        \centering
        \includegraphics[width=\linewidth]{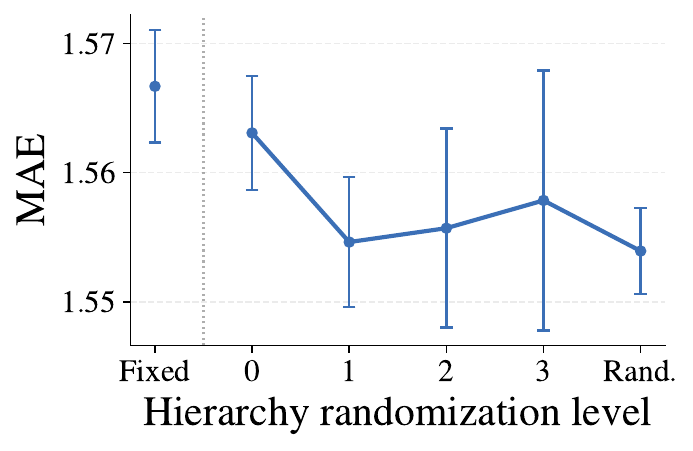}
        \caption{Variable $N_1$, performance at $L_0$.}
        \label{fig:progressive_same_val_test_bay_five_seeds}
    \end{subfigure}
    \hfill
    \begin{subfigure}[b]{0.325\textwidth}
        \centering
        \includegraphics[width=\linewidth]{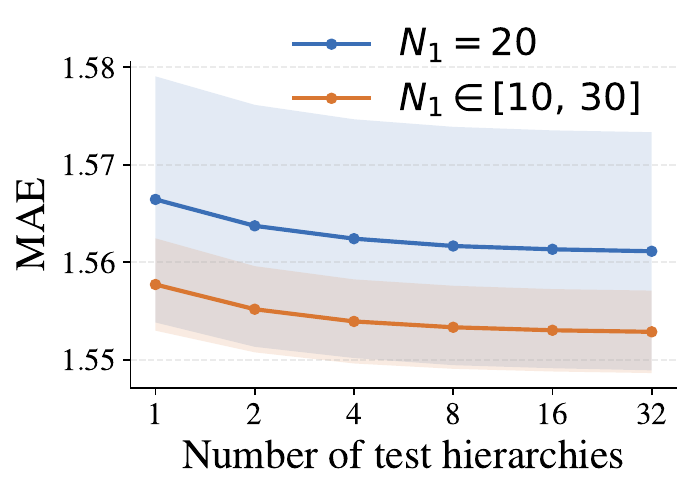}
        \caption{Test-time averaging on $L_0$.}
        \label{fig:inference_hierarchy_ensemble_bay}
    \end{subfigure}
    
    \par\vspace{2mm}
    \centering
    \begin{subfigure}[b]{0.325\textwidth}
        \centering
        \includegraphics[width=\linewidth]{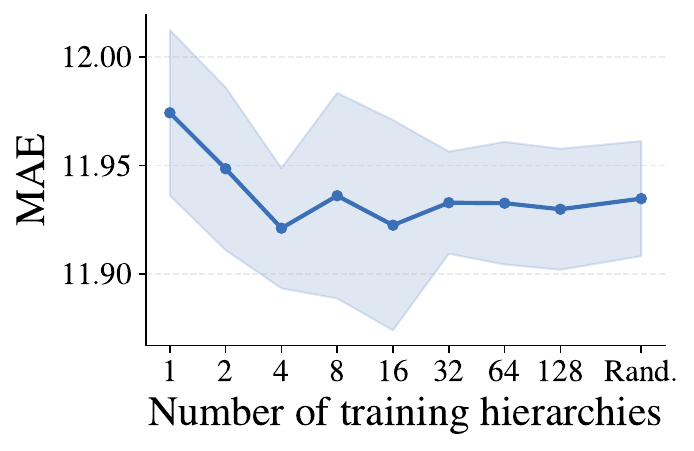}
        \caption{Fixed $N_1$, performance at $L_0$.}
        \label{fig:hierarchy_count_same_val_test_air_five_seeds}
    \end{subfigure}
    \hfill
    \begin{subfigure}[b]{0.325\textwidth}
        \centering
        \includegraphics[width=\linewidth]{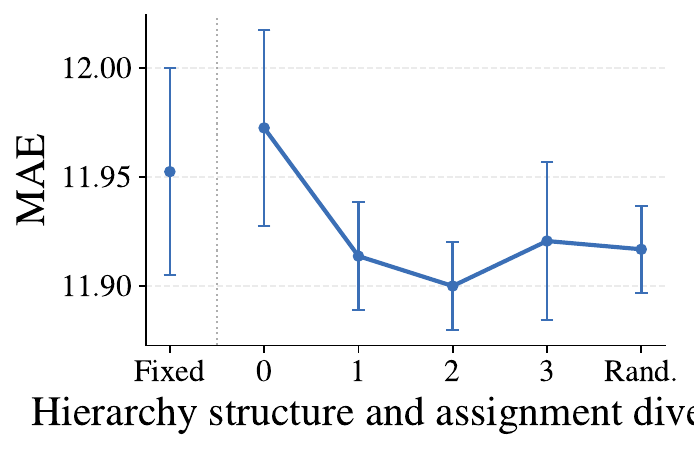}
        \caption{Variable $N_1$, performance at $L_0$.}
        \label{fig:progressive_same_val_test_air_five_seeds}
    \end{subfigure}
    \hfill
    \begin{subfigure}[b]{0.325\textwidth}
        \centering
        \includegraphics[width=\linewidth]{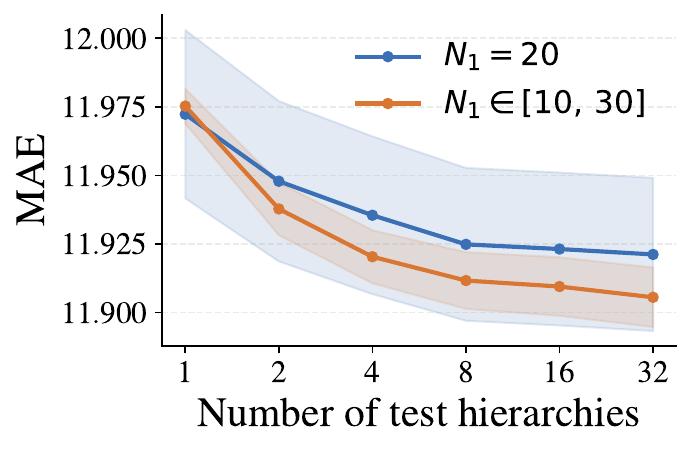}
        \caption{Test-time averaging on $L_0$.}
        \label{fig:inference_hierarchy_ensemble_air}
    \end{subfigure}
    
    \par\vspace{2mm}
    \centering
    \begin{subfigure}[b]{0.325\textwidth}
        \centering
        \includegraphics[width=\linewidth]{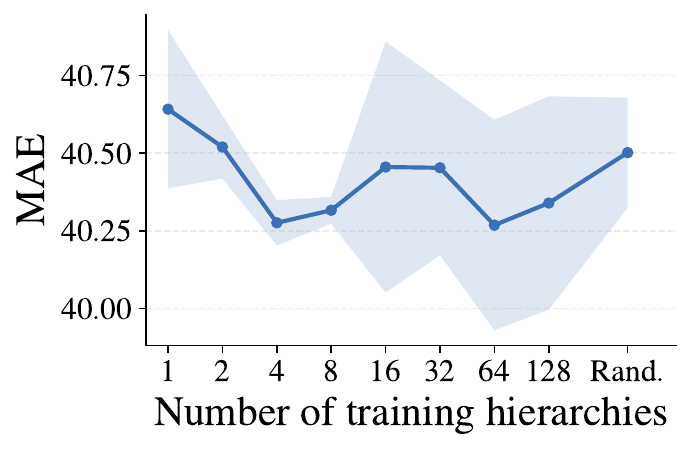}
        \caption{Fixed $N_1$, performance at $L_0$.}
        \label{fig:hierarchy_count_same_val_test_engrad_five_seeds}
    \end{subfigure}
    \hfill
    \begin{subfigure}[b]{0.325\textwidth}
        \centering
        \includegraphics[width=\linewidth]{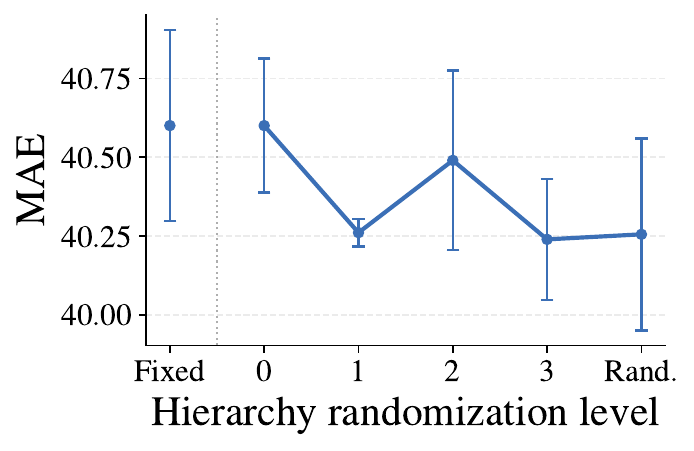}
        \caption{Variable $N_1$, performance at $L_0$.}
        \label{fig:progressive_same_val_test_engrad_five_seeds}
    \end{subfigure}
    \hfill
    \begin{subfigure}[b]{0.325\textwidth}
        \centering
        \includegraphics[width=\linewidth]{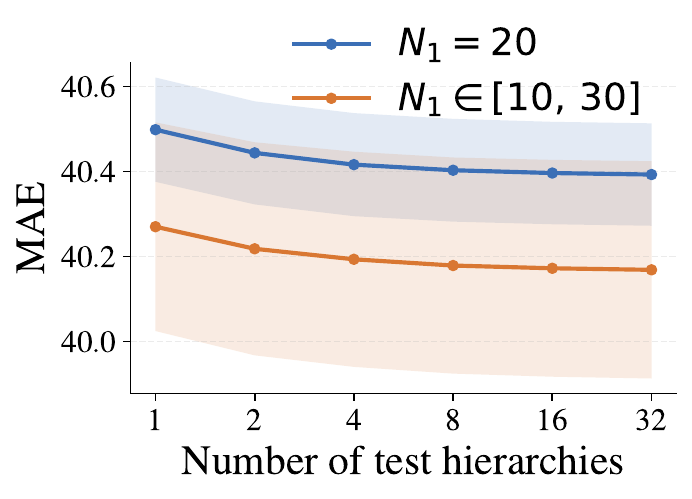}
        \caption{Test-time averaging on $L_0$.}
        \label{fig:inference_hierarchy_ensemble_engrad}
    \end{subfigure}
    
    \caption{Bottom-level performance ($L_0$) on \gls{bay} (top row), \gls{air} (middle row), and \gls{engrad} (bottom row). (a) varies the number of training hierarchies with fixed $N_1=20$; (b) increases the diversity of both node-to-aggregate assignments and the number of intermediate aggregates $N_1$; (c) varies the number of test-hierarchy forecasts being averaged, comparing fixed $N_1=20$ with $N_1\in\{10,\ldots,30\}$.}
    \label{fig:hierarchy_variability_datasets}
\end{figure}

The hierarchy used in our experiments has a pyramidal structure, with the original bottom-level nodes, a set of intermediate aggregates, and a single global node. Each bottom-level node is assigned to one intermediate node. Changing either these assignments or the number $N_1$ of intermediate nodes therefore produces a different aggregation structure. 
For the sensitivity analysis, we generally construct the intermediate aggregates by applying spectral clustering to the input graph. For \gls{air} only, we instead apply K-means to the station coordinates. We make this choice because the spectral partitions of the \gls{air} graph are particularly unbalanced, comprising a few very large aggregates and several aggregates containing only a small number of stations.
We provide below details on the sensitivity analysis presented in \autoref{sec:experiments}. Additionally, in \autoref{fig:hierarchy_variability_higher_levels_datasets}, \autoref{fig:hierarchy_variability_datasets}, we report further plots analogous to \autoref{fig:hierarchy_variability_higher_levels} and \autoref{fig:hierarchy_variability} on different datasets. 
The decreasing trends are consistent with those observed in \autoref{sec:experiments}, although varying the hierarchy structure is more beneficial on some datasets while varying node assignments is more beneficial on others.

\begin{figure}[!t]
    \centering
    \captionsetup[subfigure]{font=footnotesize,skip=2pt}

    \begin{subfigure}[!t]{0.35\textwidth}
        \centering
        \includegraphics[width=\linewidth]{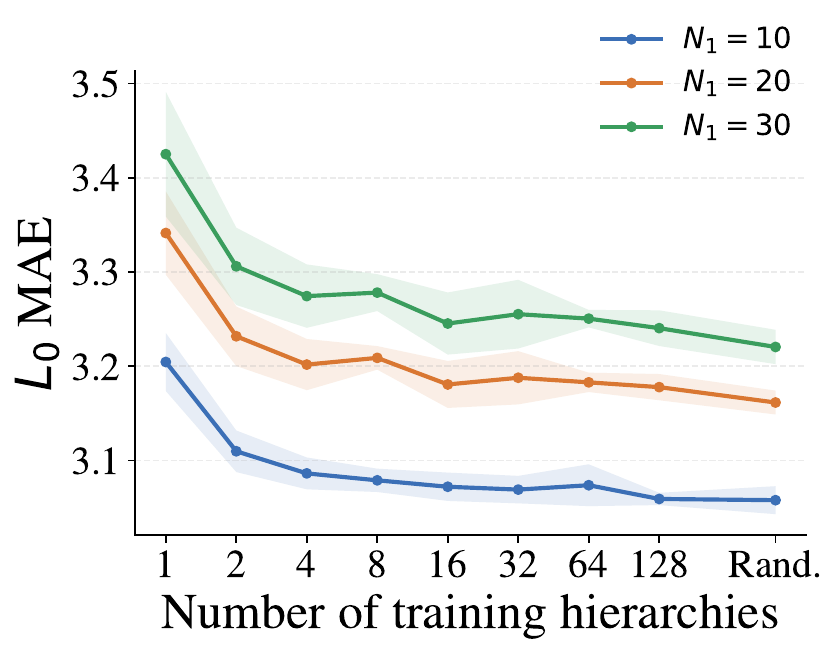}
        \caption{Fixed $N_1$, performance at $L_0$.}
        \label{fig:hierarchy_count_same_val_test_la_five_seeds_shadow_L0}
    \end{subfigure}%
    \hspace{0.15\textwidth}%
    \begin{subfigure}[!t]{0.35\textwidth}
        \centering
        \includegraphics[width=\linewidth]{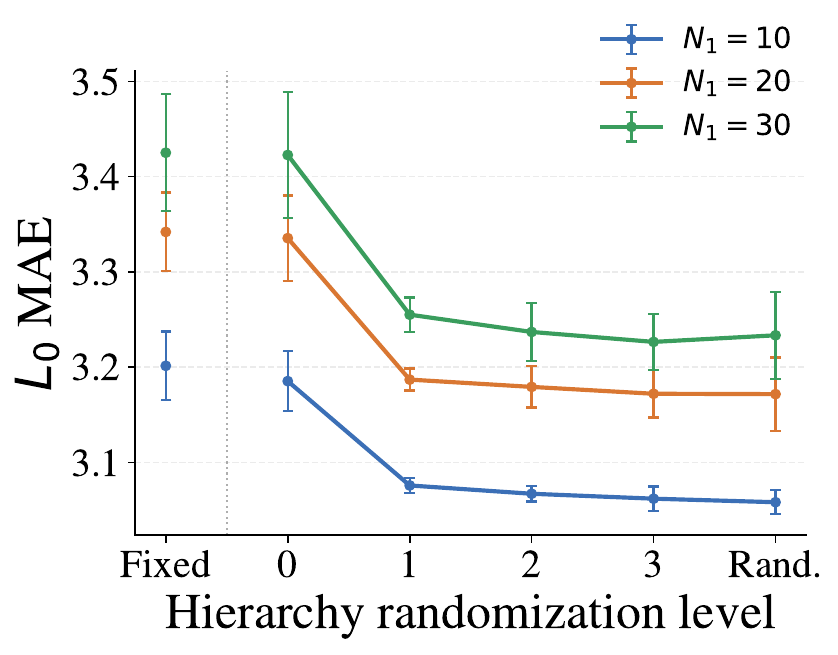}
        \caption{Variable $N_1$, performance at $L_0$.}
        \label{fig:progressive_same_val_test_la_five_seeds_L0}
    \end{subfigure}
    
    \par\vspace{2mm}
    \centering
    \begin{subfigure}[!t]{0.35\textwidth}
        \centering
        \includegraphics[width=\linewidth]{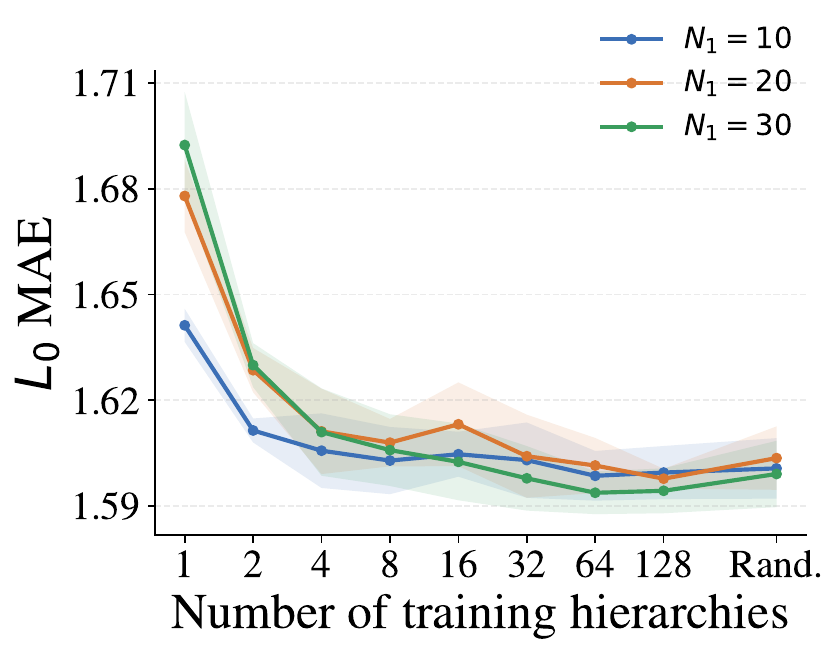}
        \caption{Fixed $N_1$, performance at $L_0$.}
        \label{fig:hierarchy_count_same_val_test_bay_five_seeds_shadow_L0}
    \end{subfigure}%
    \hspace{0.15\textwidth}%
    \begin{subfigure}[!t]{0.35\textwidth}
        \centering
        \includegraphics[width=\linewidth]{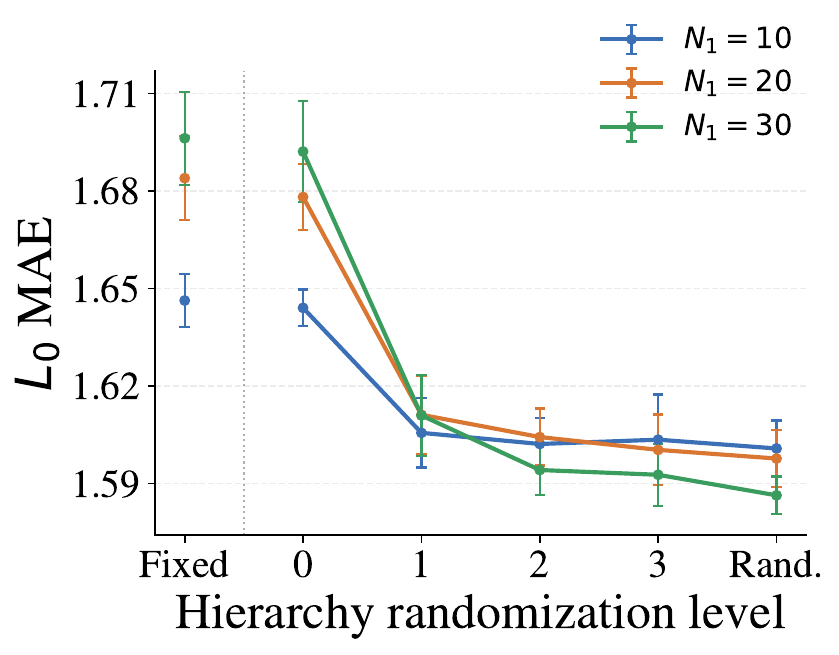}
        \caption{Variable $N_1$, performance at $L_0$.}
        \label{fig:progressive_same_val_test_bay_five_seeds_L0}
    \end{subfigure}
    
    \par\vspace{2mm}
    \centering
    \begin{subfigure}[!t]{0.35\textwidth}
        \centering
        \includegraphics[width=\linewidth]{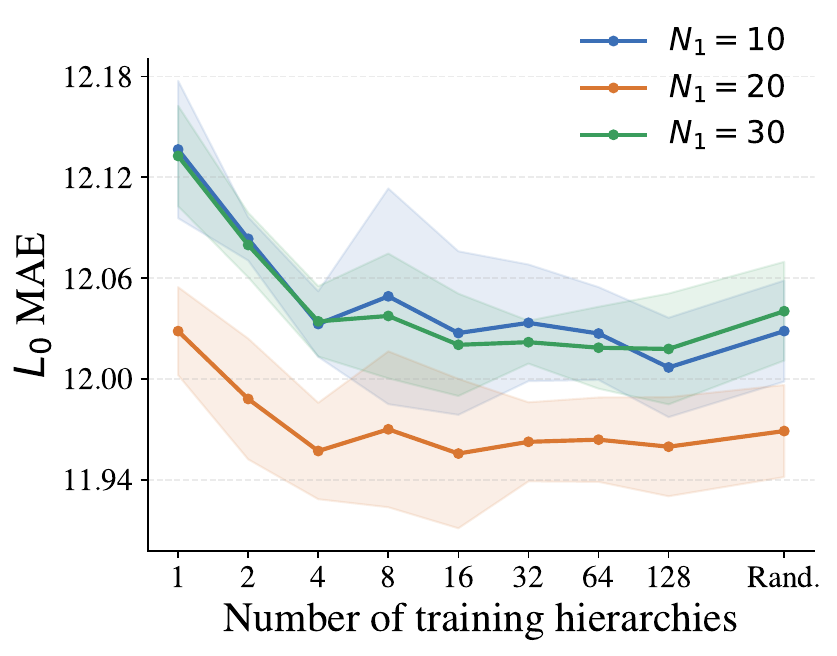}
        \caption{Fixed $N_1$, performance at $L_0$.}
        \label{fig:hierarchy_count_same_val_test_air_five_seeds_shadow_L0}
    \end{subfigure}%
    \hspace{0.15\textwidth}%
    \begin{subfigure}[!t]{0.35\textwidth}
        \centering
        \includegraphics[width=\linewidth]{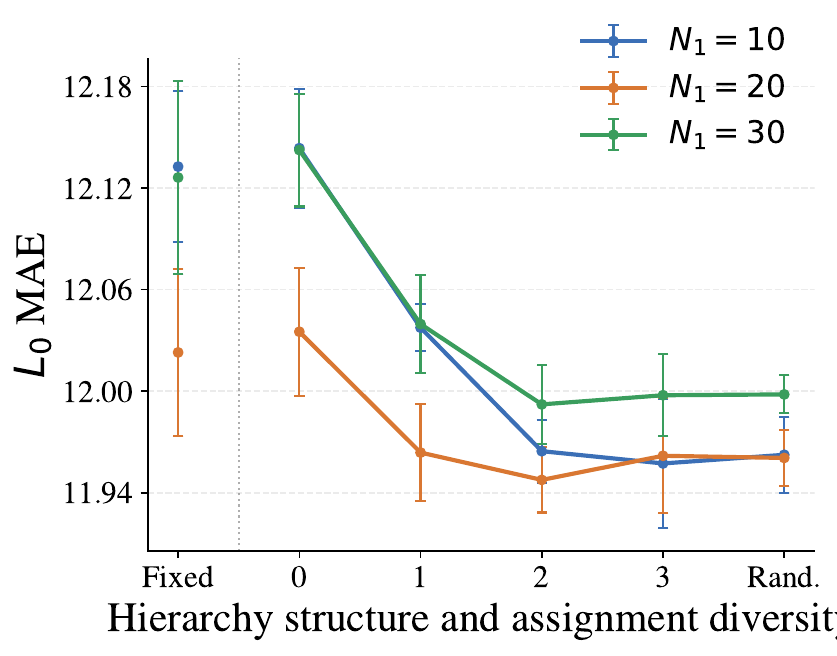}
        \caption{Variable $N_1$, performance at $L_0$.}
        \label{fig:progressive_same_val_test_air_five_seeds_L0}
    \end{subfigure}
    
    \par\vspace{2mm}
    \centering
    \begin{subfigure}[!t]{0.35\textwidth}
        \centering
        \includegraphics[width=\linewidth]{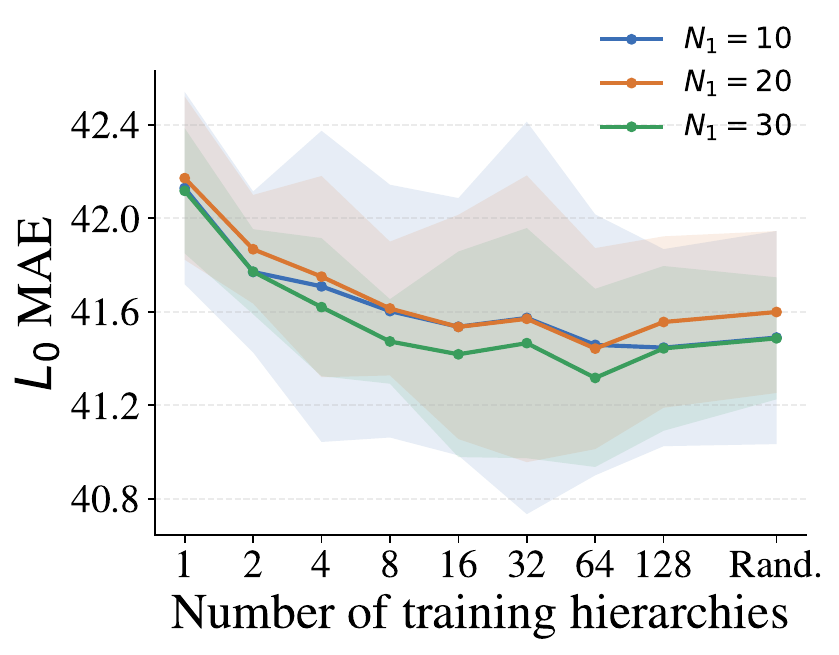}
        \caption{Fixed $N_1$, performance at $L_0$.}
        \label{fig:hierarchy_count_same_val_test_engrad_five_seeds_shadow_L0}
    \end{subfigure}%
    \hspace{0.15\textwidth}%
    \begin{subfigure}[!t]{0.35\textwidth}
        \centering
        \includegraphics[width=\linewidth]{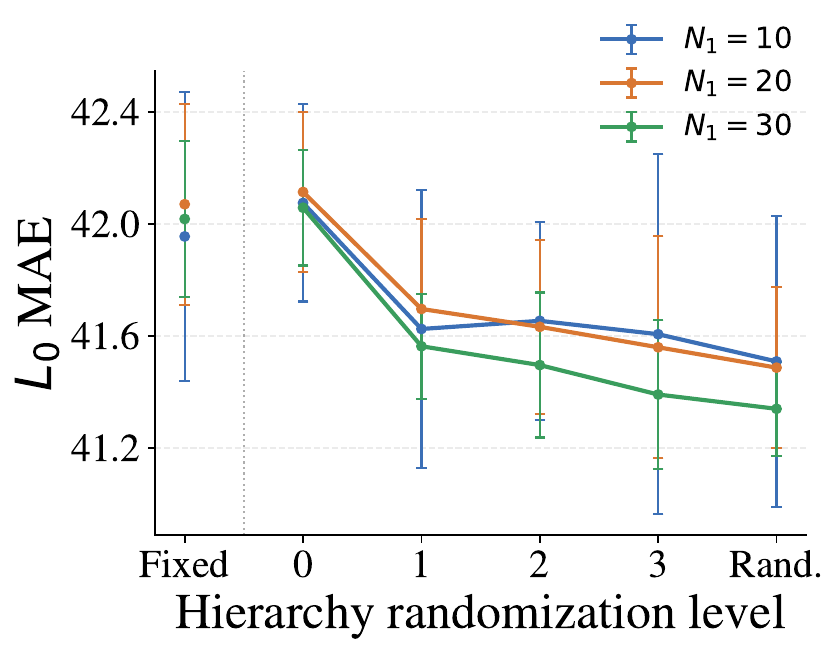}
        \caption{Variable $N_1$, performance at $L_0$.}
        \label{fig:progressive_same_val_test_engrad_five_seeds_L0}
    \end{subfigure}
    \caption{Bottom-level performance ($L_0$) on \gls{la} (first row), \gls{bay} (second row), \gls{air} (third row), and \gls{engrad} (fourth row) \textit{without inference-time hierarchy averaging} (\autoref{eq:hierarchy_averaging}). The first column increases the number of training hierarchies while keeping $N_1=20$ and varying node-to-aggregate assignments. The second column instead progressively increases training-hierarchy diversity by varying both the assignments and $N_1$. Each curve evaluates the same set of trained models using common test hierarchies with $N_1\in\{10,20,30\}$.}
    \label{fig:training_fixed_hier_L0}
\end{figure}

\paragraph{Changing node assignments.}
In \autoref{fig:hierarchy_count_same_val_test_la_five_seeds}, we isolate the effect of changing node assignments by fixing the number of intermediate nodes to $N_1=20$. Consequently, all hierarchies have the same cardinality and differ only in how bottom-level nodes are assigned to intermediate groups. 
All conditions follow the same evaluation protocol. At each validation batch, predictions are averaged across three hierarchies with $N_1=20$. 
At each test batch, predictions are averaged across ten hierarchies with $N_1=20$. The plot reports the test ensemble \gls{mae}, averaged across five training seeds, and the shaded region denotes the standard deviation.
In \autoref{fig:training_fixed_hier_L0}, we additionally evaluate the models separately on three fixed hierarchies with $N_1\in\{10,20,30\}$ and report bottom-level ($L_0$) performance. Each hierarchy determines input aggregation, hierarchical message passing, and reconciliation. We do not perform averaging in this case. Evaluating all models on the same fixed hierarchies without averaging allows us to assess the effect of training with different node assignments independently of inference-time averaging. The results show that training with multiple assignments can also improve predictions when evaluated on hierarchies with 10 and 30 intermediate nodes, although these cardinalities are not encountered during training. These results are consistent with those obtained at the aggregate levels in \autoref{sec:experiments}.

\paragraph{Changing node assignments and hierarchy structure.}
In \autoref{fig:progressive_same_val_test_la_five_seeds}, we progressively vary both the assignments of bottom-level nodes and the number of intermediate nodes.
All compared models use the same evaluation protocol, except for \textit{Fixed}. At each validation batch, predictions are averaged across three hierarchies with $N_1\in\{10,20,30\}$. At each test batch, predictions are averaged across ten hierarchies, with two independently sampled hierarchies for each $N_1\in\{10,15,20,25,30\}$. 
In \autoref{fig:training_fixed_hier_L0}, we also evaluate these models separately on the same three fixed hierarchies with $N_1\in\{10,20,30\}$ used above. These retests report \gls{mae} at the bottom level ($L_0$) for one hierarchy at a time, without applying the average in \autoref{eq:hierarchy_averaging}. As above, evaluating all models on the same fixed hierarchies without averaging allows us to assess the effect of training with different hierarchy distributions independently of inference-time averaging. Also here, the decreasing trends are consistent with those obtained in \autoref{sec:experiments}.

\subsection{Computational scalability}
\begin{figure}[t]
    \centering
    \begin{subfigure}[!t]{0.42\textwidth}
        \centering
        \includegraphics[width=\linewidth]{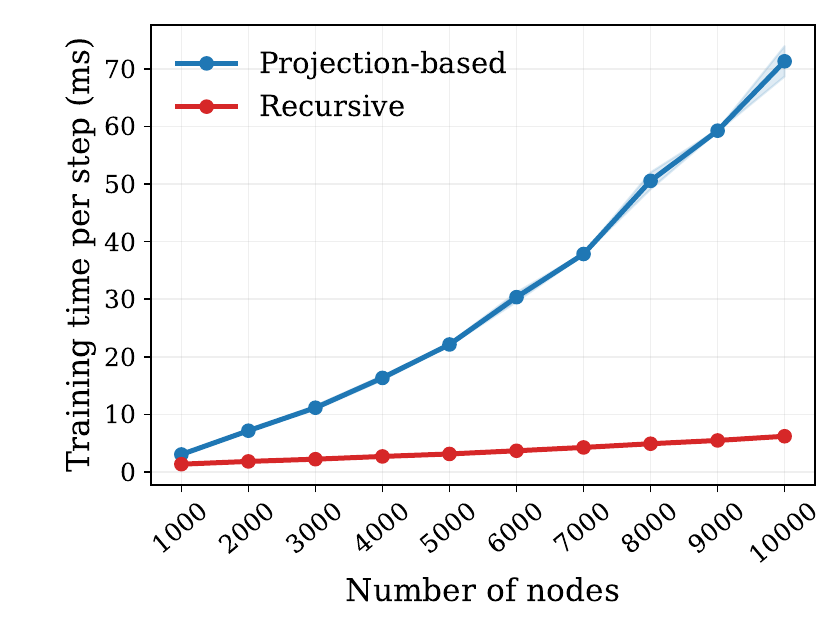}
    \end{subfigure}
    \hfill
    \begin{subfigure}[!t]{0.42\textwidth}
        \centering
        \includegraphics[width=\linewidth]{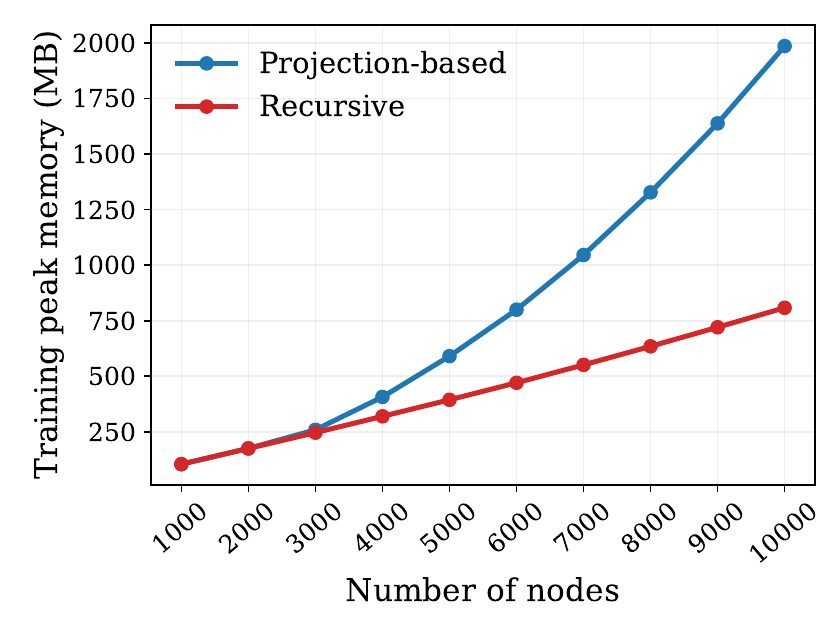}
    \end{subfigure}
    \caption{\textit{Training} scalability using projection-based reconciliation and our recursive procedure. The panels report training time (left) and peak GPU memory usage (right).}
    \label{fig:training_scaling_cuda_time_peak_memory}
\end{figure}

\begin{figure}[t]
    \centering
    \begin{subfigure}[t]{0.42\textwidth}
        \centering
        \includegraphics[width=\linewidth]{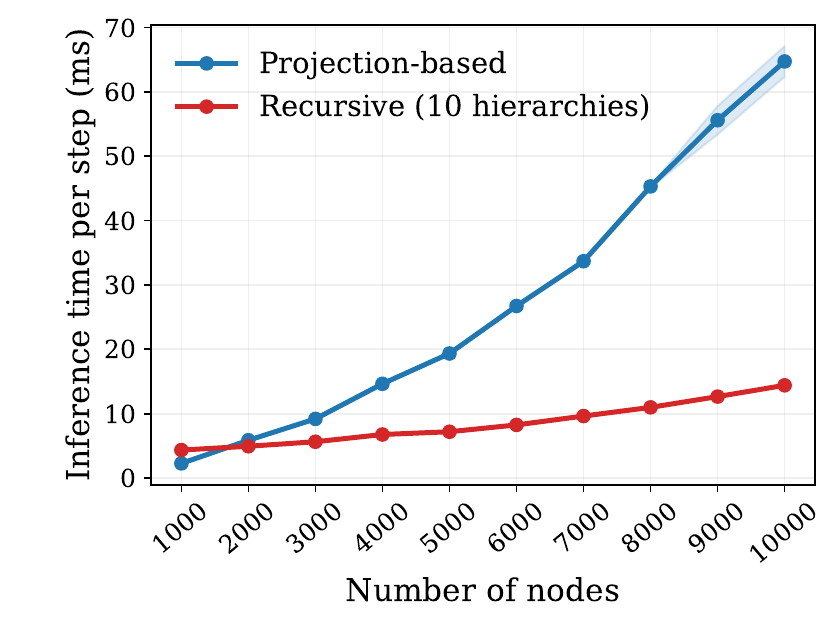}
    \end{subfigure}
    \hfill
    \begin{subfigure}[t]{0.42\textwidth}
        \centering
        \includegraphics[width=\linewidth]{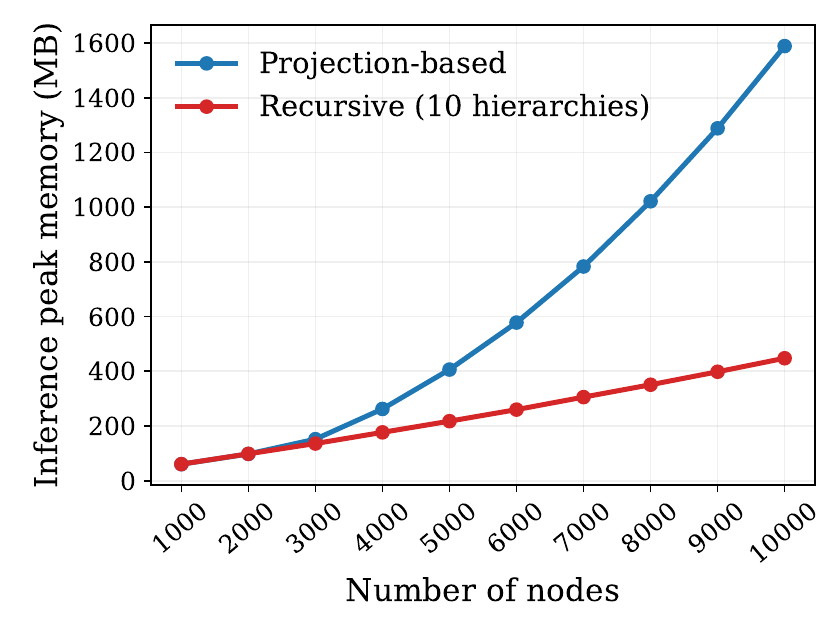}
    \end{subfigure}
    \caption{Comparison of inference scalability between projection-based reconciliation using one hierarchy and our recursive procedure using ten hierarchies. The panels report inference time (left) and peak GPU memory usage (right).}
    \label{fig:inference_scaling_cuda_time_peak_memory_10_hier}
\end{figure}

\begin{figure}[t]
    \centering
    \begin{subfigure}[!t]{0.42\textwidth}
        \centering
        \includegraphics[width=\linewidth]{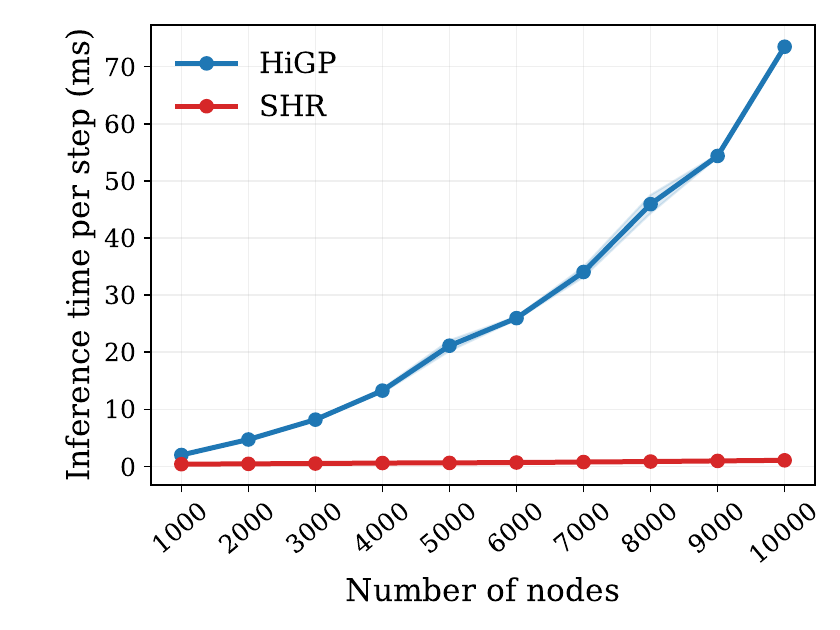}
    \end{subfigure}
    \hfill
    \begin{subfigure}[!t]{0.42\textwidth}
        \centering
        \includegraphics[width=\linewidth]{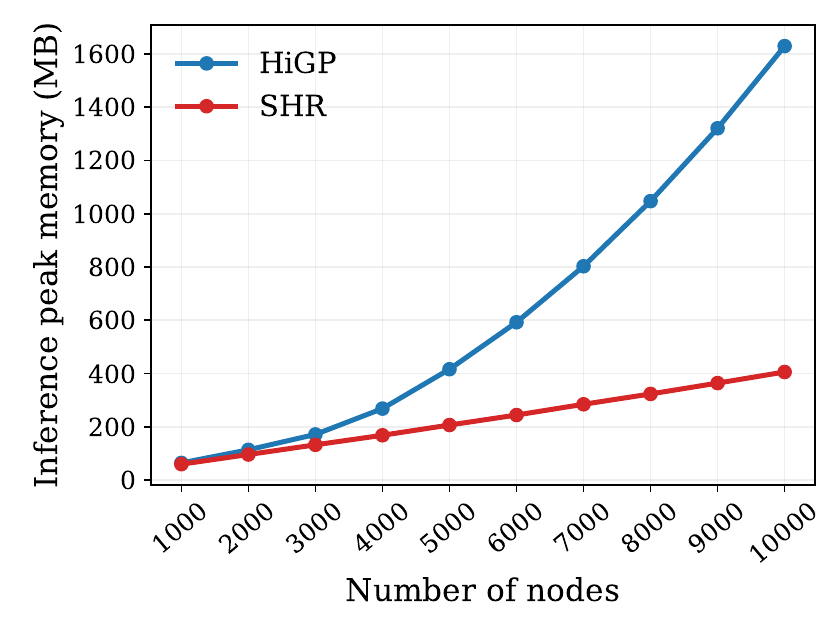}
    \end{subfigure}
    \caption{Inference scalability of \gls{higp} and \gls{shr} as the number of graph nodes increases. The panels report inference time (left) and peak GPU memory usage (right).}
    \label{fig:comparison_peak_memory_higp_shr}
\end{figure}

The experiments on computational scalability were performed on a synthetic graph made of sparse ring-connected communities linked by bridge edges. 
We measured the CUDA execution time of a single inference step using PyTorch Profiler \citep{paske2019pytorch} and the peak GPU memory allocated during the step using \textit{torch.cuda.max\_memory\_allocated}. 
All scalability experiments were conducted on a single NVIDIA RTX PRO 6000 Blackwell Max-Q Workstation Edition GPU with 96 GB of memory, FP32 precision, and a batch size of one. 
In~\autoref{fig:training_scaling_cuda_time_peak_memory}, we extend the analysis to training and observe scaling trends similar to those reported for inference in \autoref{sec:experiments}.
In~\autoref{fig:inference_scaling_cuda_time_peak_memory_10_hier}, instead, we evaluate inference scalability using ten hierarchies. Our recursive procedure is applied separately to each hierarchy, and the resulting coherent forecasts are averaged as defined in \autoref{eq:hierarchy_averaging}, whereas projection-based reconciliation processes only one hierarchy. This experiment assesses whether the scalability advantage reported in the main text is preserved when our method processes multiple hierarchies at inference time, compared with a less efficient reconciliation procedure applied to a single hierarchy. The results show that the advantage persists despite the additional hierarchies processed by our method.
Finally, in \autoref{fig:comparison_peak_memory_higp_shr}, we also account for the cost of building the hierarchy in \gls{higp} and sampling a hierarchy in \gls{shr}. Accordingly, we profile \gls{higp} using projection-based reconciliation and \gls{shr} using our recursive procedure. The resulting scaling trends remain consistent with the other experiments, suggesting that the procedure used to obtain coherent forecasts largely determines the differences in computational scalability observed in the plots.

\subsection{Inductive setting}

We evaluate the performance of \gls{shr} on held-out graphs from \gls{gla}. 
\autoref{tab:largest_inductive_step} reports additional forecasting results at horizon steps 3, 6, and 12. Both variants of \gls{shr} outperform the fixed-hierarchy baseline at all three horizons. In this setting, random hierarchy sampling yields lower mean errors than \gls{kmis}. A possible explanation is that, even with a fixed $k$, \gls{kmis} can generate substantially different numbers of aggregate nodes depending on graph topology. In the inductive setting, this can lead to differences in hierarchy size between training and unseen graphs. By contrast, the random sampling strategy constrains the number of aggregates to $\{10,\ldots,30\}$ and produces groups of approximately equal size within each graph, during both training and testing. This difference in the hierarchy distributions should be considered when interpreting the performance of the two \gls{shr} variants.

\begin{table}[ht]
\caption{\gls{mae} performance on \gls{gla} at selected forecasting horizons. \textbf{Bold} and \underline{underlined} values denote the best and second-best mean performance, respectively.}
\label{tab:largest_inductive_step}

\vspace{0.1in}
\centering
\resizebox{0.6\textwidth}{!}{
\begin{tabular}{l|r|rrr}
\toprule
Model & \multicolumn{4}{c}{\gls{gla}} \\
\cmidrule(lr){2-2}\cmidrule(lr){3-5}
 & Window & @3 & @6 & @12 \\
\midrule
Fixed hier. & 96 & 18.851 {\tiny $\pm\,.119$} & 22.934 {\tiny $\pm\,.255$} & 27.166 {\tiny $\pm\,.171$} \\
\midrule
\gls{shr} (rand.) & 96 & \textbf{17.980 {\tiny $\pm\,.119$}} & \textbf{21.652 {\tiny $\pm\,.112$}} & \textbf{25.699 {\tiny $\pm\,.029$}} \\
\gls{shr} & 96 & \underline{18.136 {\tiny $\pm\,.056$}} & \underline{21.972 {\tiny $\pm\,.144$}} & \underline{25.711 {\tiny $\pm\,.024$}} \\
\bottomrule
\end{tabular}
}
\end{table}

\section{Implementation Details} 

We implement our code in Python~\citep{rossum2009python} and use the following libraries: 
\begin{itemize}
    \item PyTorch~\citep{paske2019pytorch}; 
    \item PyTorch Geometric~\citep{fey2019fast}; 
    \item PyTorch Lightning~\citep{Falcon_PyTorch_Lightning_2019};   
    \item Torch Spatiotemporal~\citep{Cini_Torch_Spatiotemporal_2022}; 
    \item Hydra~\citep{Yadan2019Hydra}; 
    \item Numpy~\citep{harris2020array}.
\end{itemize}

\end{document}